\documentclass[sigconf, nonacm, authorversion]{acmart}

\usepackage{subcaption}
\usepackage{colortbl}
\usepackage{xspace}

\usepackage{soul}
\sethlcolor{black}
\makeatletter

\AtBeginDocument{%
  }

\newcommand{\floodrisk}{r_c}
\newcommand{\ourmethod}{\textsc{BayFlood}\xspace}
\newcommand{\numNewResidentsCovered}{113,738}
\newcommand{\numNonThreeOneOneCovered}{433,079}
\newcommand{\numNonFloodNetCovered}{927,908}
\newcommand{\numNonDEPCovered}{293,095}

\newcommand{\decStormImages}{158,555}
\newcommand{\janStormImages}{331,034}
\newcommand{\californiaStormImages}{24,383}

\usepackage{placeins}

\begin{document}

\title{Bayesian Modeling of Zero-Shot Classifications\\ for Urban Flood Detection\\}

\author{Matt Franchi}
\email{mattfranchi@cs.cornell.edu}
\orcid{}
\affiliation{
    \institution{Cornell University, Cornell Tech}
    \city{New York}
    \state{New York}
    \country{USA}
}

\author{Nikhil Garg}
\email{ngarg@cornell.edu}
\orcid{}
\affiliation{
    \institution{Cornell Tech}
    \city{New York}
    \state{New York}
    \country{USA}
}

\author{Wendy Ju}
\email{wendyju@cornell.edu}
\orcid{}
\affiliation{
    \institution{Jacobs Technion-Cornell Institute, Cornell Tech}
    \city{New York}
    \state{New York}
    \country{USA}
}

\author{Emma Pierson}
\email{emmapierson@berkeley.edu}
\orcid{}
\affiliation{
    \institution{University of California - Berkeley}
    \city{Berkeley}
    \state{California}
    \country{USA}
}

\renewcommand{\shortauthors}{Franchi et al.}

\begin{abstract}
Street scene datasets, collected from Street View or dashboard cameras, offer a promising means of detecting urban objects and incidents like street flooding. However, a major challenge in using these datasets is their lack of reliable labels: there are myriad types of incidents, many types occur rarely, and ground-truth measures of where incidents occur are lacking. Here, we propose \ourmethod, a two-stage approach which circumvents this difficulty. First, we perform \emph{zero-shot} classification of where incidents occur using a pretrained vision-language model (VLM). Second, we fit a spatial Bayesian model on the VLM classifications. The zero-shot approach avoids the need to annotate large training sets, and the Bayesian model provides frequent desiderata in urban settings --- principled measures of uncertainty, smoothing across locations, and incorporation of external data like stormwater accumulation zones. We comprehensively validate this two-stage approach, showing that VLMs provide strong zero-shot signal for floods across multiple cities and time periods, the Bayesian model improves out-of-sample prediction relative to baseline methods, and our inferred flood risk correlates with known external predictors of risk. Having validated our approach, we show it can be used to improve urban flood detection: our analysis reveals \numNewResidentsCovered~people who are at high risk of flooding overlooked by current methods, identifies demographic biases in existing methods, and suggests locations for new flood sensors. More broadly, our results showcase how Bayesian modeling of zero-shot LM annotations represents a promising paradigm because it avoids the need to collect large labeled datasets and leverages the power of foundation models while providing the expressiveness and uncertainty quantification of Bayesian models. 
\end{abstract}

\received{2025.}

\maketitle

\section{Introduction}
Street scene datasets, derived from dashboard cameras ("dashcams") or Street View data, offer an unparalleled view into urban life. They have been used to count urban objects, including trees \cite{berland_google_2017, li_quantifying_2018, branson_google_2018}, traffic signs \cite{campbell_detecting_2019}, curb ramps \cite{hara_tohme_2014} and manholes \cite{vishnani_manhole_2020}; measure inequality in policing and surveillance \cite{franchi_detecting_2023,sheng2021surveilling}; estimate demographics \cite{gebru_using_2017}, pedestrian counts \cite{yin_big_2015}, safe infrastructure \cite{rundle_using_2011}, navigability \cite{yin_measuring_2016, franchi_matt_robotability_2025} and gentrification \cite{ilic_deep_2019}; and measure neighborhood changes over time \cite{naik_computer_2017}.  

However, a major challenge in using street scene data is acquiring \emph{large labeled datasets} with which to train computer vision models to detect objects of interest \cite{rundle_using_2011}. This is challenging for several reasons. First, there are \emph{myriad types} of objects and incidents we might wish to detect. Past work has studied hundreds of types of urban incidents~\cite{balachandarusing}; thousands of types of vehicles \cite{gebru_using_2017}; and hundreds of types of trees \cite{berland_google_2017}. Each of these types necessitates its own labels. A second challenge is that many types of urban phenomena \emph{appear rarely} --- for example, street flooding occurs infrequently and does not affect most streets --- creating a class imbalance problem which can make it challenging to curate sufficient positive examples with which to train and evaluate a model. A final obstacle is that ground-truth for many urban phenomena is difficult to obtain: for example, resident reporting systems identifying where urban problems occur are noisy and have demographic biases \cite{agostini_bayesian_2024, kontokosta_bias_2021, mclafferty_placing_2020, balachandarusing,liu2024redesigning,liu2024quantifying}.

Because obtaining large labeled datasets is challenging, an appealing solution is to instead perform \emph{zero-shot} classification using pretrained vision-language models (VLMs): for example, by prompting the model to classify whether a street image shows flooding. While this avoids the need for large labeled datasets, on its own it is inadequate for several reasons. First, we would like to reliably estimate \emph{uncertainty} in flood risk estimates due to, for example, error in the zero-shot classifications or small samples of images in a given area. Second, we would like to incorporate \emph{prior knowledge} to inform our estimates: for example, if we believe flooding is spatially correlated, we might wish to smooth over spatially adjacent areas. Third, we might want to incorporate \emph{external data} --- for example, known predictors of flood risk --- to improve our estimates. 

We thus propose a two-stage approach, \ourmethod, which leverages the strengths of modern VLMs and uses classical Bayesian methods to overcome their limitations. In the first stage, we use VLMs to perform zero-shot classification of where incidents occur. We then randomly select a small number of classified positives and classified negatives and obtain ground-truth annotations. In the second stage, we fit a spatial Bayesian model on the model classifications $\hat y$ and ground-truth annotations $y$. This model naturally accommodates the desiderata mentioned above: it provides principled estimates of multiple sources of uncertainty; captures prior knowledge that ground-truth should be spatially correlated across adjacent locations; and incorporates external data.

We illustrate the benefits of \ourmethod by applying it to detect urban floods, leveraging a unique dataset of 1.4 million street images from multiple days and cities when flooding occurred. We conduct four validations of \ourmethod, showing that (1) VLM classifications provide strong signal for flood risk across multiple cities and time periods; (2) our Bayesian model improves out-of-sample prediction relative to baseline methods; (3) our approach can be applied even with very few ground-truth labels; and (4) our inferred flood risk correlates with known external predictors of flood risk. Having validated \ourmethod, we show that our flood detections can usefully augment three methods of flood risk prediction used by urban decision-makers --- resident (311) flooding reports; flood sensors; and stormwater accumulation zones. Specifically, \ourmethod reveals flooded areas missed by each of these methods and affecting \numNewResidentsCovered\space people; highlights biases in resident reports; and suggests locations for new flood sensors, which we are providing to the organization which places the sensors as part of our ongoing conversations.

Overall, we propose a general two-stage approach for detecting objects and incidents in unlabeled street scene datasets which leverages the complementary strengths of VLMs and Bayesian models. Our approach avoids the need to collect large labeled datasets by relying on the zero-shot classification abilities of VLMs, while providing the expressiveness and uncertainty estimation of Bayesian models. This approach is applicable to the many settings in which street scene datasets are useful, including in computational social science, urban sensing, and public health \cite{biljecki_street_2021, see_review_2019, rzotkiewicz_systematic_2018}. More broadly, our approach highlights the benefits of combining modern foundation models with classical statistical methods which use their annotations as input --- an idea which has powerful applications in many other settings~\cite{angelopoulos2023prediction,cherian2024large,gligoric2024can,shanmugam2025evaluating}.%

\section{Related work}
We discuss four lines of related work: vision models applied to street images; Bayesian modeling of urban phenomena; modeling language model predictions using classical statistical methods; and flood detection. 

\subsection{Vision models applied to street images}
Domain-specific vision models have been trained using supervised learning to detect specific objects (including street trees \cite{berland_google_2017, li_quantifying_2018, branson_google_2018}, traffic signs \cite{campbell_detecting_2019}, curb ramps \cite{hara_tohme_2014}, manholes \cite{vishnani_manhole_2020}, pedestrians~\cite{yin_big_2015}, and vehicles~\cite{gebru_using_2017,franchi_detecting_2023}) and predict neighborhood characteristics~\cite{rundle_using_2011,ilic_deep_2019}. Earlier works relied on Google Street View \cite{biljecki_street_2021, rzotkiewicz_systematic_2018, vandeviver_applying_2014}, and more recent works explore temporally \textit{denser} street imagery \cite{franchi_detecting_2023, franchi_towards_2024} that permits analyses of more short-horizon phenomena, like vehicle deployment rates or spatiotemporal trends in pedestrian traffic.

More recent models like CLIP have made zero-shot image classification possible \cite{radford_learning_2021}. Now, large labeled datasets are no longer necessary for supervised learning. %
CLIP, and models derived from it, have been applied to diverse tasks including geo-location (determining the location of an image anywhere on Earth)~\cite{haas_learning_2023}; extracting building attributes~\cite{pan_zero-shot_2024}; estimating land use~\cite{wu_mixed_2023}; and inferring urban functions~\cite{huang_zero-shot_2024}. Subsequent to CLIP, a new generation of vision-language models (VLMs), including API-accessible models like GPT-4V \cite{openai_gpt-4_2024} and Gemini Pro \cite{team_gemini_2024} as well as open-source models like Cambrian-1 \cite{tong_cambrian-1_2024} and DeepSeek's Janus Pro \cite{chen_janus-pro_2025}, offer higher generalizability and performance \cite{zhang_gpt-4vision_2023}. While much work in the urban science domain relies on earlier CLIP-based models, in our work we rely on this newer generation of models (specifically, Cambrian).

\subsection{Bayesian modeling of urban phenomena}

Bayesian methods have been applied in many settings relevant to urban life, including book transfer in public libraries \cite{liu_identifying_2024}, crowdsourced citizen reporting systems \cite{agostini_bayesian_2024,laufer2022end}, policing~\cite{pierson2020large,pierson2018fast,simoiu2017problem}, and healthcare and public health~\cite{balachandar2023domain,chiang2024learning,pierson2020assessing}. In general, Bayesian models are widely employed due to their expressiveness, ability to incorporate prior knowledge, and principled quantification of uncertainty~\cite{gelman1995bayesian}, all properties we leverage in our present work. 

\subsection{Modeling LM predictions using classical statistical methods}

A rich prior literature has showcased the benefits of modeling predictions from VLMs, LLMs, or other machine learning models using classical statistical methods. For example, ~\cite{gligoric2024can} develops a method for modeling LLM predictions and confidence indicators to strategically select which human annotations are needed and provide valid confidence intervals. ~\cite{angelopoulos2023prediction} models machine learning predictions in combination with other experimental data and develops a method for producing valid confidence intervals. ~\cite{shanmugam2025evaluating} models the joint distribution of machine learning predictions and ground-truth labels to estimate model performance. A number of papers develop conformal prediction methods to provide principled statistical performance guarantees for LLM outputs~\cite{cherian2024large,mohri2024language,quach2023conformal}. These works highlight the benefits of modeling predictions from VLMs, LLMs, and other models using classical statistical methods, motivating our two-stage approach. 

\subsection{Flood detection}
We apply our method to \emph{flood detection} both because it is an important problem and because rich, newly-available data exists to validate our method.  Flooding endangers lives, causes serious economic impacts, and is growing worse with climate change \cite{newman_global_2023, hinkel_coastal_2014, brody_rising_2007, desmet_evaluating_2018}. Since 2000, flooding has affected 1.6 billion people globally, caused at least 651 billion USD in damages, and led to more than 130,000 fatalities \cite{un_office_for_disaster_risk_reduction_human_2020, ritchie_natural_2022, devitt_flood_2023}. Flooding costs the United States on the order of $180$ billion dollars yearly \cite{us_congress_joint_economic_committee_jec_2024}. Here, we study flooding impacts in urban environments, which can be catastrophic: for example, one day of rainfall in New York City on September 29, 2023 -- depicted in our dashcam dataset -- caused over $\$100$ million dollars in damage \cite{aon_weekly_2023}. 

The globally-significant impacts of flooding have motivated a rich prior literature on near-realtime flood detection.  One approach is crowdsourced detection, or `social sensing' \cite{arthur_social_2018}, of flooding through social media posts~\cite{chaudhary_flood-water_2019, witherow_floodwater_2019, alizadeh_human-centered_2022, wang_hyper-resolution_2018, geetha_detection_2017, narayanan_novel_2014, park_computer_2021}; these approaches interface with the larger idea of citizen science, which is used as an important component in building community resilience \cite{see_review_2019}. A related literature explores flood detection from citizen reporting services like 311 \cite{agostini_bayesian_2024, rainey_using_2021, agonafir_machine_2022, agonafir_understanding_2022}. Sensors for flood detection have also been deployed: e.g., the FloodNet project installs physical ultrasonic sensors \cite{ceferino_developing_2023, silverman_making_2022, mousa_flash_2016} above intersections in flood-prone areas that are capable of monitoring flooding in real-time \cite{mydlarz_floodnet_2024}. Machine learning methods are often deployed to process raw meteorological data from sensors or satellite data (see \cite{mosavi_flood_2018} for a comprehensive review.) For example, \cite{mauerman_high-quality_2022} have developed a Bayesian latent variable model to predict seasonal floods in Bangladesh via the fusion of two satellite data streams. Predictive flooding models \cite{nearing_global_2024} have been developed to cover 100 countries, 700 million people, and increase the effective lead-time for extreme river flooding events to 7 days~\cite{yossi_matias_how_2024, nearing_global_2024}.

Closest to our own work is the literature which seeks to detect flooding from image data. This includes work on real-time flood detection via networks of CCTV surveillance cameras and other live camera feeds \cite{bhola_flood_2018, jafari_real-time_2021, liang_v-floodnet_2023, hao_estimating_2022, narayanan_novel_2014}. Satellite images have also been explored as a medium for flood detection, when paired with machine learning and computer vision methods \cite{brakenridge_flood_2003, munawar_remote_2022, mason_near_2012, klemas_remote_2014}. Our work differs from this literature because it relies on temporally-dense dashcam data for flood detection, which has not been previously explored and, more fundamentally, develops a general and novel two-stage methodology for urban object and incident detection which is applicable in many settings beyond flood detection.

\section{Data}

We now describe the data used in this paper. The primary input to our method, which we describe in \S \ref{sec:dashcam_data}, is dashcam images of public street scenes \cite{franchi_towards_2024}. We supplement this data and validate our flood risk estimates with additional data sources we describe in \S \ref{sec:additional_datasets}, including government-produced open datasets \cite{janssen_benefits_2012}, physical flooding sensors \cite{mydlarz_floodnet_2024}, and predictive stormwater accumulation maps \cite{nyc_department_of_environmental_protection_2024_nodate}.

\subsection{Dashcam data}
\label{sec:dashcam_data}
Consumer, vehicle-mounted dashcams, known for their utility in safety and protective liability, provide spatially and temporally dense image data, capturing the urban \textit{streetscape}. Relative to prior datasets, like Google Street View, dashcam datasets offer much higher temporal density, rendering them superior for analyzing short-horizon events like flooding; in contrast, the gap in time between consecutive images in Google Street View can be as large as 7 years \cite{kim_examination_2023}.

\begin{figure}
    \includegraphics[width=0.4\textwidth]{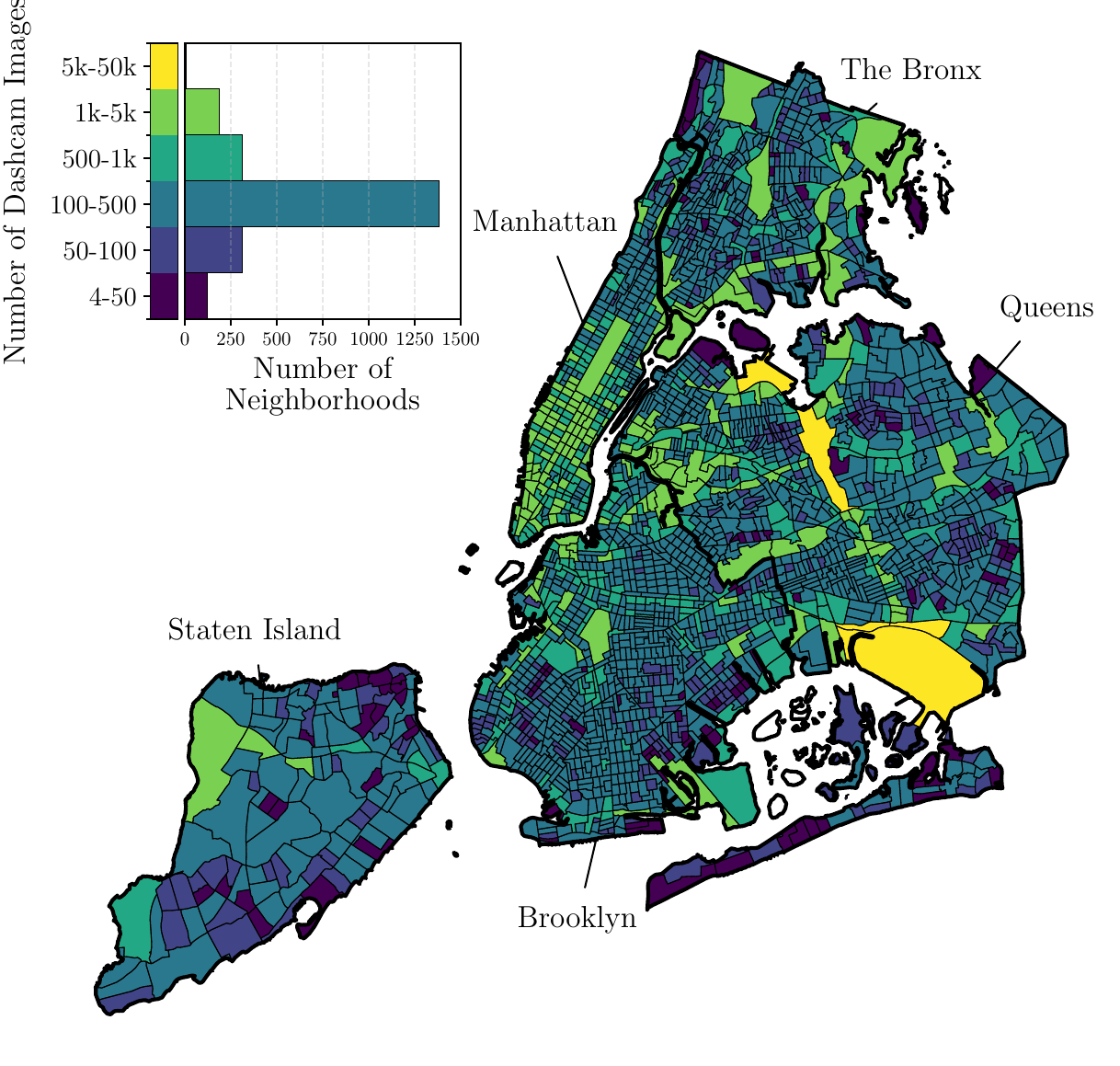}
    \caption{Spatial distribution of dashcam images in our primary analysis dataset in New York City. Most Census tracts have one hundred to five hundred images.}
    \label{fig:ct-dashcam-dist}
\end{figure}

\begin{figure*}[h!]
    \renewcommand*{\arraystretch}{0.5}
    \centering
    \begin{subfigure}[b]{0.66\textwidth}
        \centering
        \begin{tabular}{cc}
            \includegraphics[trim={0 1.3cm 0 2.1cm}, clip, width=0.5\textwidth]{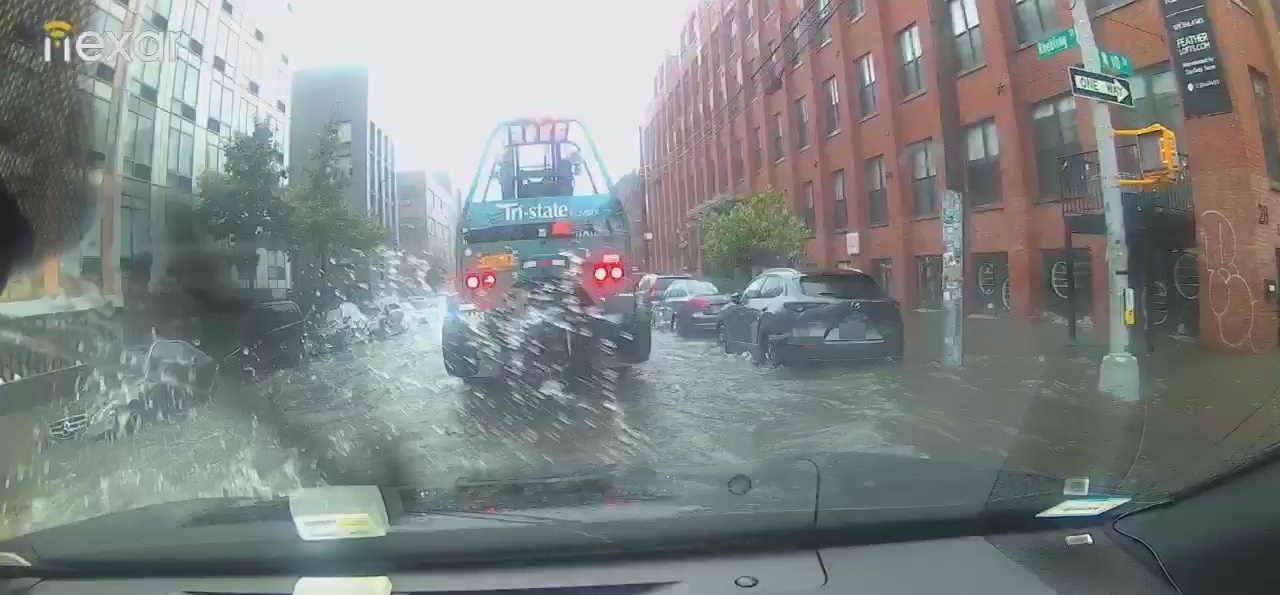} & 
            \includegraphics[trim={0 1.5cm 0 3cm}, clip, width=0.5\textwidth]{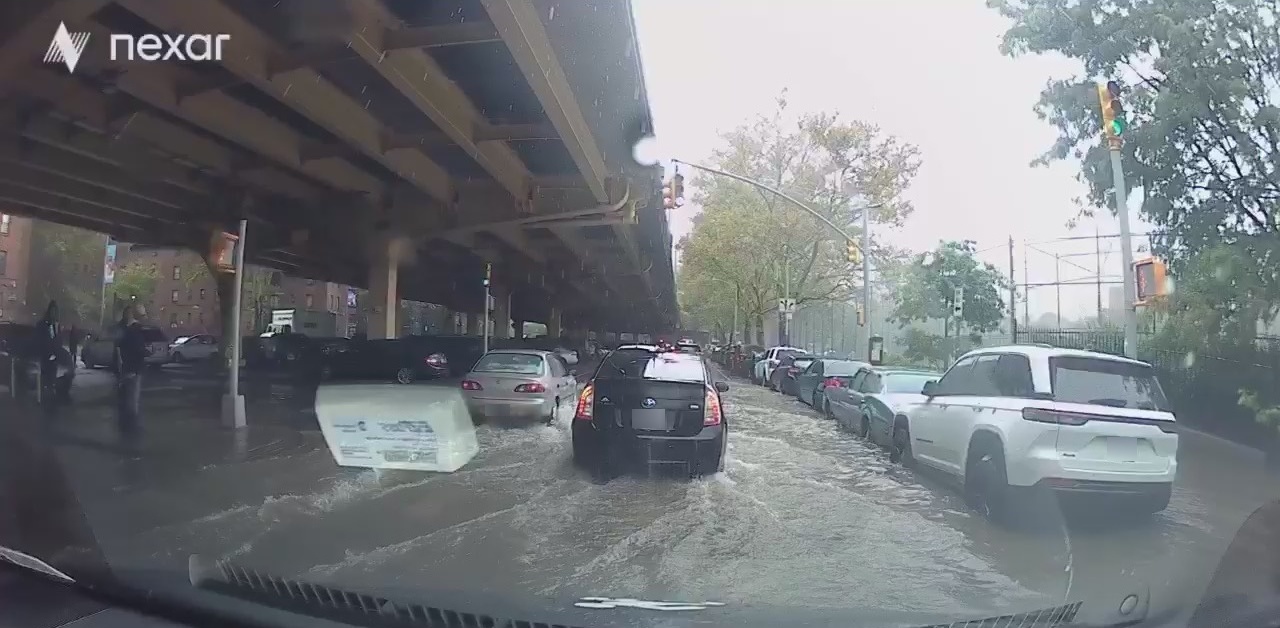} \\
            \includegraphics[trim={0 1.3cm 0 4cm}, clip, width=0.5\textwidth]{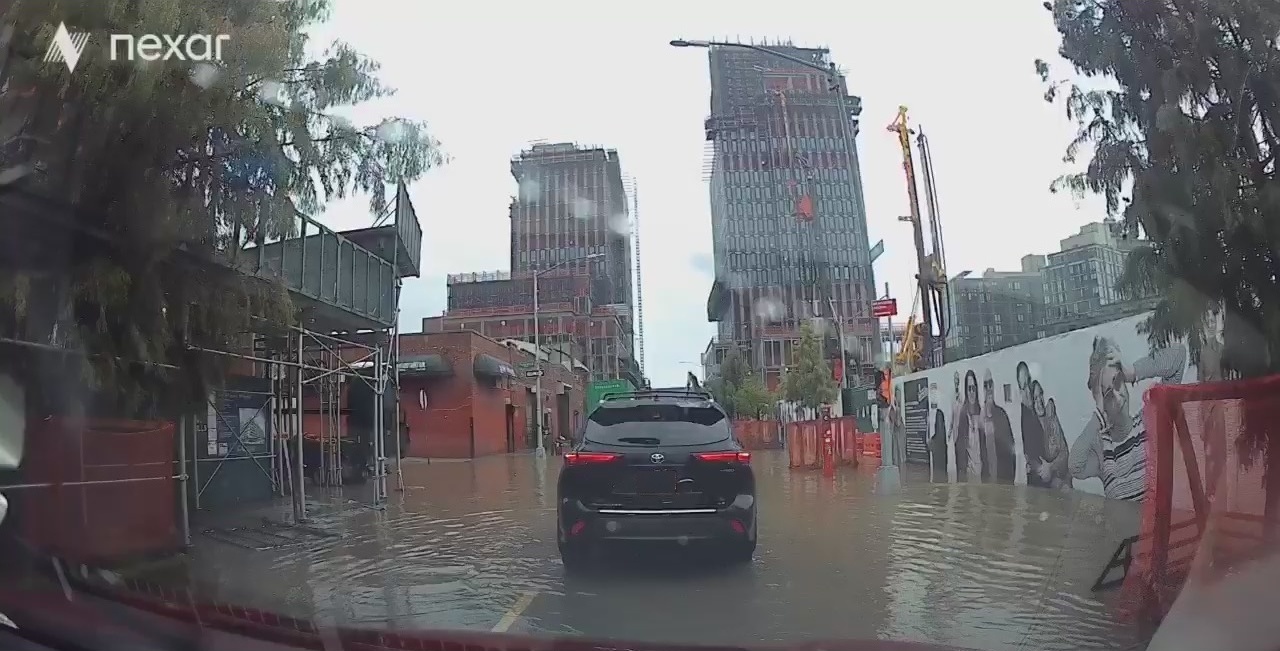} & 
            \includegraphics[trim={0 1.3cm 0 3.6cm}, clip, width=0.5\textwidth]{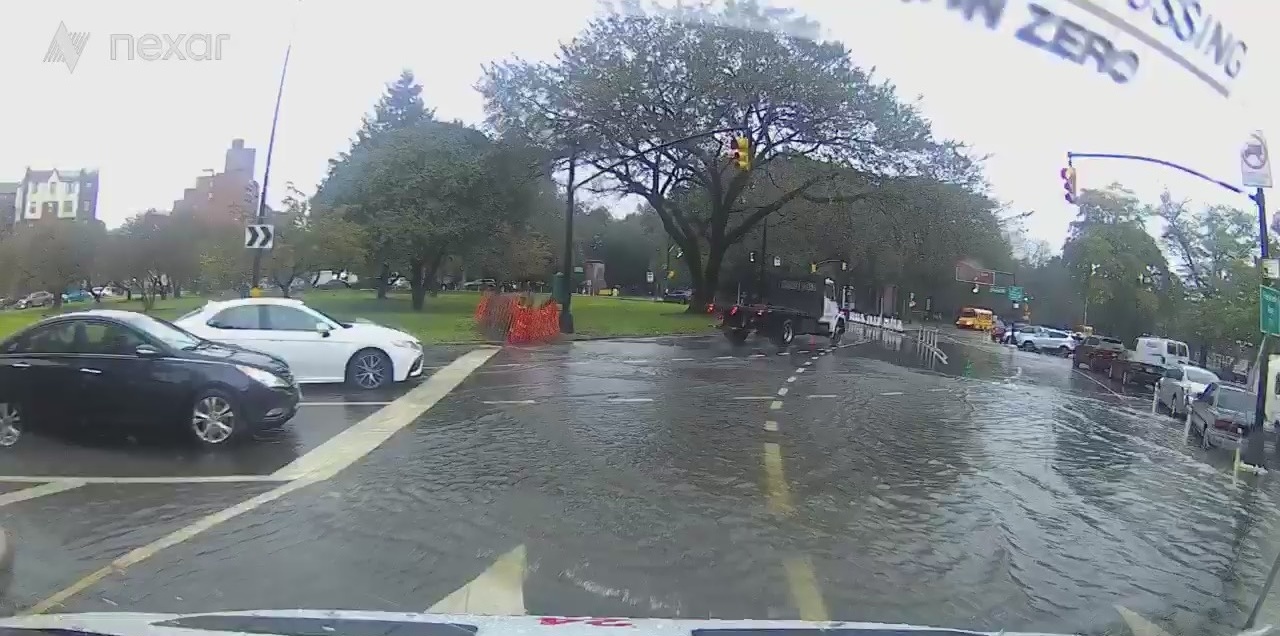}
        \end{tabular}
        \caption{True Positives}
        \label{fig:tp}
    \end{subfigure}
    \hfill
    \begin{subfigure}[b]{0.32\textwidth}
        \centering
        \begin{tabular}{c}
            \includegraphics[trim={0cm 0cm 0cm 2cm}, clip, width=\textwidth]{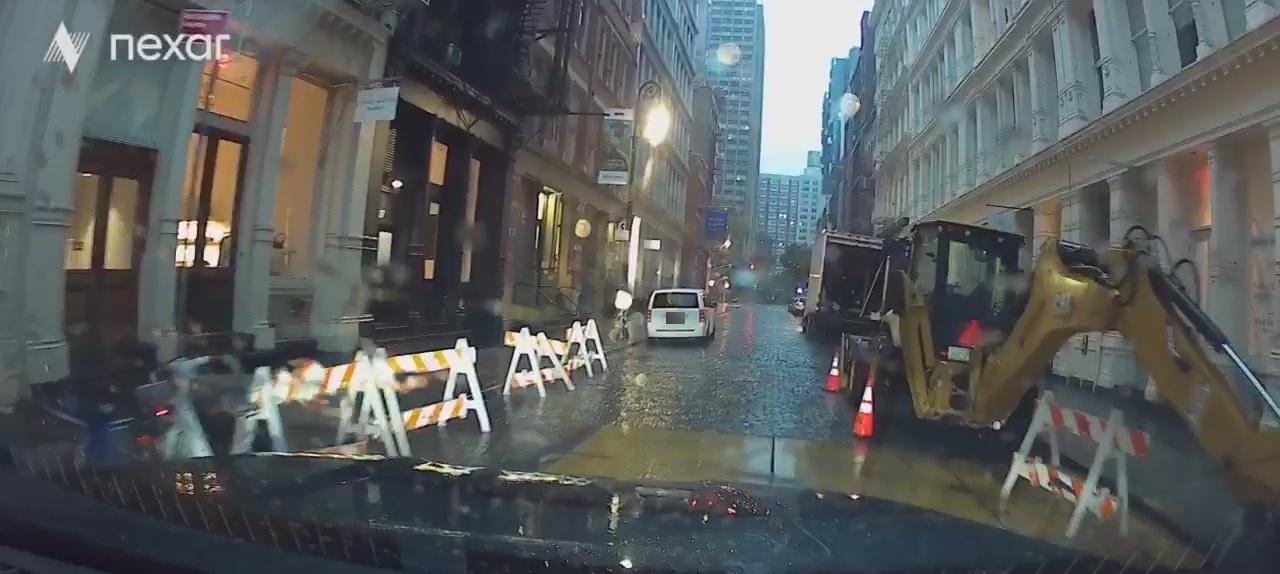} \\
            \includegraphics[trim={0cm 0cm 0cm 1cm}, clip, width=\textwidth]{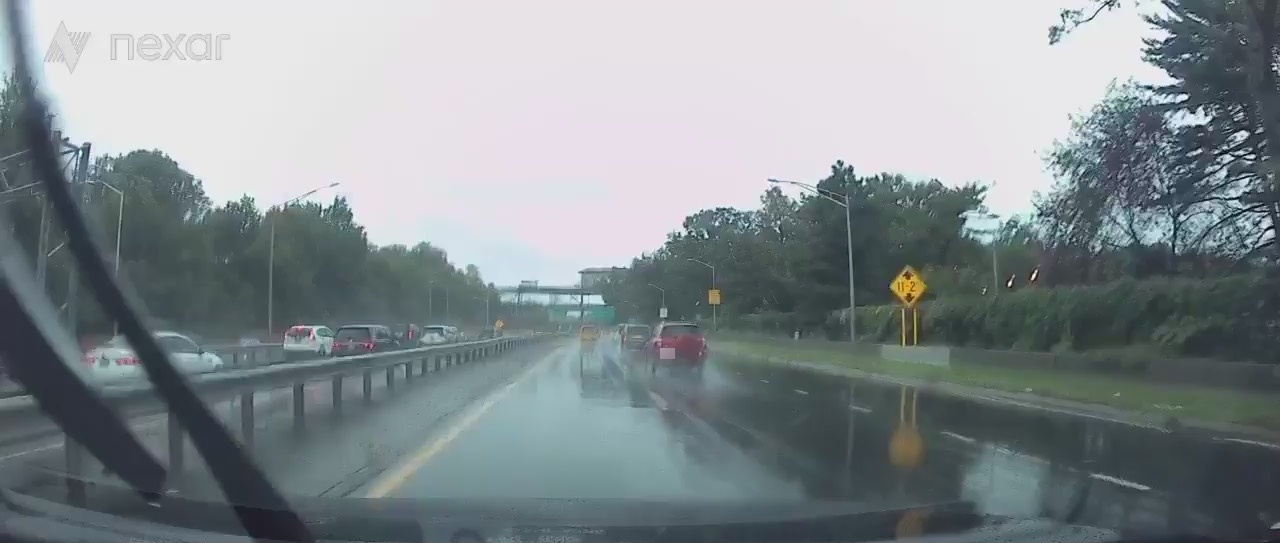}
        \end{tabular}
        \caption{False Positives}
        \label{fig:fp}
    \end{subfigure}

    \caption{Representative true and false positive flood classifications from the VLM. False negatives are extremely rare due to the low prevalence of flooding in the dataset.}
    \label{fig:detections}
\end{figure*}

Our dashcam dataset is provided by Nexar, whose data has been widely used in prior work \cite{dadashova_detecting_2021, franchi_detecting_2023, franchi_towards_2024, shapira_fingerprinting_2024, chowdhury_designing_2024, chowdhury_towards_2021}. Nexar images are 1280 $\times$ 720 pixels and are captured from cameras affixed to the windshield of actively-driving vehicles, mostly those of ridesharing\footnote{Ridesharing refers to services that offer on-demand passenger pickup and dropoff at a chosen destination; companies that offer ridesharing include Uber and Lyft.} drivers. We develop custom tooling to cull imagery of interest from Nexar's data moat. Our primary dataset consists of 926,212 images from a storm in New York City on September 29, 2023 which caused widespread flooding. We additionally validate our approach on Nexar images from three other days: \decStormImages\space images from New York City on December 17-18, 2023; \janStormImages\space images from New York City on January 9-10, 2024; and \californiaStormImages\space images from the San Francisco area \cite{noauthor_february_2025} on February 10, 2024. All dates are chosen because they coincide with storms which caused known flooding events. In our primary analysis dataset on September 9, 2023, the median Census tract contains 220 images, and only 5.2\% of tracts have fewer than 50 images.\footnote{\emph{Census tracts} are fine-grained geographic areas within the United States with 4,000 inhabitants on average. New York City has 2,327 Census tracts.} \autoref{fig:ct-dashcam-dist} depicts the spatial distribution of images, and \autoref{fig:detections} provides examples of representative images. An important strength of our analysis is that we develop and validate our flood detection method using more than a million spatiotemporally granular dashcam images across flood events from multiple dates and cities. To our knowledge, due to the rarity of floods and the difficulty of collecting temporally dense street scene data at the spatial scale of a city, a dataset with these characteristics has not been previously used. 

\paragraph{Ethics} Our use of this dataset has been previously deemed not human subjects research by our institution's IRB, as our data depicts public street scenes and we do not analyze pedestrians. We are committed to ethical use of our data, and our data provider maintains a high data anonymization standard of blurring pedestrians, license plates, and dashboards prior to us having any access.  The data provider additionally blacks out the top and bottom of each image to remove any personally-identifiable information from the driver that may appear on the vehicle dashboard.

\setlength{\tabcolsep}{0.1em}
\setlength{\tabcolsep}{0.5em}

\subsection{Additional datasets}
\label{sec:additional_datasets}
We make use of the following external datasets relevant to flood risk to contextualize and validate our flood risk predictions.

\paragraph{311 reports}
Crowdsourced resident reporting systems like NYC311 have emerged as important indicators of infrastructural problems, including street flooding. In a typical 311 system, residents have the ability to submit reports of non-emergency problems (via app, internet, or phone) which are then routed to the appropriate city agency for remediation \cite{minkoff_nyc_2016}. During the September 29, 2023 storm in New York City, for example, there were 2,171 calls made to 311 pertaining to flooding-related issues (see \S\ref{si:datasets} for the list of issues we define as flooding-related). While 311 provides valuable information on potential floods, and is thus a useful external validation of our model's flood detections, it is also known to contain biases due to disparities in how likely neighborhoods are to report problems~\cite{white_promises_2018, kontokosta_equity_2017,agostini_bayesian_2024, kontokosta_bias_2021,clark2013coproduction,o2015ecometrics}. We show our approach can be applied to audit these biases. %

\paragraph{Physical flooding sensors}
Physical flooding sensors are an important current source of flooding signal for cities~\cite{silverman_making_2022}. We rely on data from FloodNet \cite{mydlarz_floodnet_2024}, a New York City government-academic partnership to develop low-cost, easy-to-assemble flooding sensors and install them throughout high-risk areas. %
At the time of the flood on which we conduct our primary analysis (September 9, 2023) there were 67 active FloodNet sensors; as of December 12, 2024 (the last update), there are 253 unique FloodNet sensors placed at some point in time. We use the locations of these 253 sensors as an indicator of flood risk. Placement decisions are informed by a detailed, community-informed process \cite{floodnet_nyc_community_2025}.

\paragraph{Stormwater accumulation maps}
Stormwater accumulation maps have been developed as a useful tool for governments and citizens to facilitate flood readiness \cite{van_alphen_flood_2009, faber_superstorm_2015, rosenzweig_npcc4_2024}. We use Stormwater Flooding Maps from the New York City Department of Environmental Protection (NYC DEP)  \cite{nyc_department_of_environmental_protection_2024_nodate}, which use simulations that incorporate drainage system data and flow capacity measurements to provide estimates of (1) shallow flooding (more than 4 inches, less than 1 foot) and (2) deep and contiguous flooding (greater than 1 foot). We select a version of the map that simulates a moderate stormwater flood, which best replicates the conditions on the date our primary analysis dataset is collected.

\paragraph{Digital elevation maps (DEM)}
We utilize New York City's Digital Elevation Map \cite{nyc_opendata_1_2024} to compute basic elevation metrics for each Census tract in the city, including minimum elevation, mean elevation, and maximum elevation. The DEM was generated with one-foot granularity using LiDAR data, and is meant to ground elevation in feet above sea level, with all built surface features removed. %

\paragraph{American Community Survey (ACS) demographic data.}
We use the US Census' American Community Survey to investigate our model's population coverage and investigate biases in 311 data. Following similar work \cite{kontokosta_bias_2021, agostini_bayesian_2024, kontokosta_equity_2017}, we select datasets on total population, reported race and age \cite{Census2023ACSDP5Y2023.DP05}, household income \cite{Census2023ACSST5Y2023.S1901}, educational attainment \cite{Census2023ACSST5Y2023.S1501}, access to technology \cite{Census2023ACSST5Y2023.S2801}, and language spoken at home \cite{Census2023ACSST5Y2023.S1602}. 

\section{Method}

We now describe our method, \ourmethod. \ourmethod has two stages. First, using a VLM, we perform zero-shot classification of whether dashcam images show flooding, and annotate a small number of classified positives and classified negatives with ground-truth human labels (\S \ref{sec:methods_vlm}). Second, we use the classifier labels $\hat y$ and the ground-truth annotations $y$ as inputs to a Bayesian spatial model which smooths across adjacent areas and incorporates external flood risk features (\S \ref{sec:methods_bayesian_model}). The raw images are not used as inputs to the Bayesian model.

\subsection{Zero-shot VLM classification} 
\label{sec:methods_vlm}

We perform zero-shot classification of whether each image is flooded using the Cambrian model~\cite{nyu_visionx_cambrian-1_2024}, an open-source VLM developed in 2024 which achieved state-of-the-art results among open-source models like LLaVA-NeXT \cite{liu_llava-next_2024}, and comparable performance to the best proprietary models including GPT-4V \cite{openai_gpt-4_2024} and Gemini-Pro \cite{team_gemini_2024}. We use the 13B-parameter version of the model, and use the prompt \textit{Does this image show more than a foot of standing water?} This prompt was the most performant on the annotated inspection set we describe below, and aligns with the definition of `deep' flooding as defined by the New York City Department of Environmental Protection \cite{nyc_department_of_environmental_protection_2024_nodate}. We use the 13B-parameter version of Cambrian because it considerably outperforms the smaller 8B-parameter version (Table \ref{tab:vlm-baselines}) without introducing the significantly higher inference costs of the 34B-parameter version. Performing inference on all 926,212 images in our primary dataset takes approximately one week when distributed across 6 Nvidia RTX A6000 GPUs. The model classifies 0.2\% of images as flooded, consistent with the imbalanced nature of the dataset.

\paragraph{Measuring model performance} We assess Cambrian-1-13B's performance on our primary dataset by randomly sampling 500 images classified as positive and 500 classified as negative, and manually annotating them. 
One researcher from the team annotated all images to ensure annotation criteria were consistent. Each image was annotated as positive if it showed \emph{definite flooding}: namely, the street in front of the vehicle was visible and showed significant flooding. Ambiguous images, and ponding and other small puddles, were marked as negative. We quantify model performance by reporting the positive predictive value $p(y=1|\hat y=1)$ (i.e., the proportion of classified positives which are truly positive) and the false omission rate $p(y=1|\hat y=0)$ (i.e., the proportion of classified negatives which are truly positive). We also validate model performance on three additional dashcam image datasets from other dates and cities (\S \ref{sec:dashcam_data}).

\paragraph{Comparison to classification baselines} We compare the zero-shot classification performance of Cambrian-1-13B to that of several other VLMs: Cambrian-1-8B, CLIP, and DeepSeek Janus-Pro-7B. We also compare to a supervised learning baseline (a ResNet fine-tuned on a subset of the dataset with flooding labels). We fully describe these baselines in Appendix \ref{si:additional_vlm_details}. For all models, we compare performance using the same metrics discussed above --- namely, for each model, we estimate $p(y=1|\hat y=1)$ and $p(y=1|\hat y=0)$ by taking a random sample of its positive classifications, and a random sample of its negative classifications, and annotating with ground-truth labels.

\subsection{Bayesian modeling of VLM classifications}
\label{sec:methods_bayesian_model}

After classifying all images using the VLM, and manually annotating a small subset of the classified images, we then fit a Bayesian model on the 926,212 model classifications (positive or negative) and manual ground-truth annotations (positive, negative, or unknown). Because we only annotate 1,000 images, the vast majority of annotations are unknown. The raw images are not used as inputs to the Bayesian model. 

The purpose of the Bayesian model is to estimate the proportion of images in each Census area $c$ which are truly flooded, $p(y=1|C=c)$, while accounting for uncertainty due to classifier error and finite samples of images; smoothing across adjacent areas; and incorporating external data relevant to flood risk.

\paragraph{Observed data} Let $y$ denote the manual ground-truth annotation for each image (i.e., whether it truly flooded) and $\hat y$ its label from the VLM classifier. In each Census area, we have images of six types, depending on (a) whether the image's classifier label is positive or negative and (b) the ground-truth annotation label is positive, negative, or unknown (2 possibilities $\times$ 3 possibilities = 6 image types). Thus, our observed data for each Census area $c$ consists of a set of six numbers: the counts of images in the Census area  $n^{(c)}_{\hat y = \ell_{\hat y}, y = \ell_y}$ where the classifier label is $\ell_{\hat y} \in \{0, 1\}$ and the ground-truth label is $\ell_{y} \in \{0, 1, ?\}$. For example, the observed data for one Census area with 100 images might be ``90 images were classified negative, and have unknown ground-truth label; 9 were classified positive, and have unknown ground-truth label; and 1 was classified positive, and has a positive ground-truth label''. For each Census area $c$ we additionally observe a vector $X_c$ of flood-relevant features from the data sources described in \S \ref{sec:additional_datasets}: for example, whether the area is a flood risk zone or has any resident complaints of flooding. (Appendix \ref{sec:flooding_features_in_bayesian_model} lists features and describes feature preprocessing.)

\paragraph{Model} We summarize our model here and provide additional details in Appendix \ref{si:bayesian_modeling}. Our main quantity of interest is the probability that an image in a Census area  $c$ shows flooding, $p(y=1|C=c)$. We model this as follows:
\begin{equation*}
p(y=1|C=c)=\text{logit}^{-1}(\alpha + X_c\beta + \phi_c \cdot \sigma_\phi)
\end{equation*}
where $\alpha$ is an intercept term, $\beta$ is the feature coefficients, and $\phi_c$ is an Intrinsic Conditional Auto-Regressive (ICAR) spatial component which varies by Census area, a standard technique to capture spatially correlated phenomena like flooding~\cite{morris2019bayesian,besag1995conditional} by smoothing across adjacent areas. 

We model VLM classifier errors by introducing parameters to capture the classifier's true positive rate $\theta_{\hat y = 1 | y = 1} \triangleq p(\hat y = 1 | y = 1)$ and false positive rate $\theta_{\hat y = 1 | y = 0} \triangleq p(\hat y = 1 | y = 0)$. We assume these error rates remain constant across Census areas.%

The log likelihood (LL) of the observed data in Census area $c$ is:
\begin{align*}
    \underbrace{\sum_{\ell_{\hat y} = 0, 1}\sum_{\ell_y = 0, 1}n^{(c)}_{\hat y = \ell_{\hat y}, y = \ell_y} \text{log }p(\hat y = \ell_{\hat y}, y = \ell_y|C=c)}_{\text{LL of images \emph{with} ground-truth labels}} +  \\\underbrace{\sum_{\ell_{\hat y} = 0, 1}n^{(c)}_{\hat y = \ell_{\hat y}, y = ?} \text{log }p(\hat y = \ell_{\hat y}|C=c)}_{\text{LL of images \emph{without} ground-truth labels}}
\end{align*}

We can write $p(\hat y = \ell_{\hat y}, y = \ell_y|C=c)$ and $p(\hat y = \ell_{\hat y}|C=c)$ in terms of $p(y|C=c)$ and the error rates $\theta$, allowing us to express the LL of the observed data in terms of the model parameters: 
\begin{align*}
p(\hat y = 1, y = \ell_y|C=c) &= p(y=\ell_y|C=c) \cdot \theta_{\hat y = 1 | y = \ell_y}\\
p(\hat y = 0, y = \ell_y|C=c) &= p(y=\ell_y|C=c) \cdot \big(1 - \theta_{\hat y = 1 | y = \ell_y}\big)\\
p(\hat y = \ell_{\hat y}|C=c) &= p(\hat y = \ell_{\hat y}, y = 1|C=c) + p(\hat y = \ell_{\hat y}, y = 0|C=c)
\end{align*}

To complete the Bayesian model specification, we place weakly informative priors over all model parameters. We fit the model using Hamiltonian Monte Carlo (HMC)~\cite{neal2012mcmc,chen2014stochastic} as implemented in the probabilistic programming language Stan~\cite{carpenter2017stan}. Below, we will use $\floodrisk \triangleq p(y=1|C=c)$ as shorthand to refer to the model's inferred flood risk in a given Census tract.%

\section{Results}
We first perform four validations of \ourmethod (\S\ref{sec:results_validation}), showing that (1) the VLM classifier provides strong signal for detecting flooded images, and outperforms baselines; (2) the Bayesian modeling approach improves out-of-sample prediction relative to baselines; (3) our predictions remain robust even with very few ground-truth annotations; and (4) our inferred measures of flood risk correlate with external ground-truth markers not used in model fitting. Having validated \ourmethod, we show that it can be usefully applied to improve flood detection in New York City (\S\ref{sec:improving_flood_detection}), identifying flooded areas missed by current approaches, revealing inequities in coverage, and suggesting locations for additional flood sensors. 

\subsection{Method validation}
\label{sec:results_validation}

\subsubsection{The VLM classifier can detect flooded images}

On our primary dataset of images, the VLM classifier displays strong signal for differentiating flooded and non-flooded images. The positive predictive value, $p(y=1|\hat y=1)$, is 0.658, indicating that of the images the VLM classifies as flooded, 65.8\% are truly flooded; $p(y=1|\hat y = 0) = 0.006$, indicating that of the images the VLM classifies as not flooded, only 0.6\% are truly flooded. Put another way, if the VLM predicts an image is flooded, it is 110$\times$ more likely to be flooded. These metrics show both that the classifier clearly provides strong signal for flooding, and that it is imperfect, motivating our use of a Bayesian model to estimate its error rate and incorporate ground-truth annotations. Importantly, Table \ref{tab:vlm-baselines} additionally shows that our chosen model (Cambrian-1-13B) outperforms all classification baselines, achieving higher $p(y=1|\hat y=1)$ ($p < 0.001$, t-test), and lower but comparable $p(y=1|\hat y = 0)$ (differences not statistically significant, t-test).

\begin{table}[htb!]
\small
\begin{tabular}{p{3.8cm}p{2cm}p{2cm}}
\toprule
Method & $p(y=1|\hat y = 1)$ & $p(y=1|\hat y = 0)$ \\
\midrule
Supervised learning & 0.464 & 0.012 \\ 
CLIP & 0.224 & 0.008 \\
DeepSeek Janus-Pro-7B & 0.248 &  0.012 \\
Cambrian-1-8B & 0.152 & 0.012 \\
\textbf{Cambrian-1-13B (ours)} & \textbf{0.658} & \textbf{0.006} \\ 
\bottomrule
\end{tabular}
\caption{Comparison of our preferred VLM classifier (Cambrian-1-13B) to classification baselines on our primary dataset. Cambrian-1-13B achieves higher $p(y=1|\hat y = 1)$ than all baselines ($p < 0.001$, t-test) and comparable $p(y=1|\hat y = 0)$ to all baselines (differences not statistically significant).}
\label{tab:vlm-baselines}
\end{table}

\vspace{-12pt}

We assess how well the VLM classifier generalizes to other days and cities by measuring its performance during two other floods in New York City and an additional flood in the San Francisco Bay area (\S\ref{sec:dashcam_data}). Performance remains strong (\autoref{tab:other-days-performance}): images which are classified as flooded are at least\footnote{In our three validation datasets, we do not observe any false negatives among the images classified negative. We thus compute these numbers using an upper bound of one false negative.} 351, 406, and 72 times likelier to be flooded than images which are not across the three days.

\subsubsection{Bayesian modeling improves out-of-sample prediction} 

Having validated the first stage of \ourmethod (VLM classification of whether flooding occurs) we now validate the second (fitting a spatial Bayesian model on the classifications). Specifically, we show that our Bayesian approach improves predictions of where flooded images will occur on a held-out test set, relative to both simple heuristics (e.g., the fraction of images which are classified as positive by the VLM) and machine learning baselines. 

We perform this validation as follows. After classifying the 926,212 images in our primary dataset with the VLM, we partition them into a train set (which we use to fit the Bayesian model and the baselines on the classifications) and a test set (which we use to assess out-of-sample performance). We use three metrics to assess predictive performance on the Census tract level: (1) Pearson correlation with fraction of images in the tract which are classified flooded; (2) AUC for predicting whether the Census tract will have any classified flooded images; (3) AUC for predicting whether the Census tract will have any ground-truth annotated flooded images. To minimize the noisiness of these metrics on the test set, we reserve 70\% of the dataset for the test set. Thus, the train set for this validation consists of the VLM classifications $\hat y$, and ground-truth annotations $y$, for 30\% of the images; the test set consists of the VLM classifications and ground-truth annotations for the remaining 70\%. 

We compare to three sets of baselines. First, we compare to several \emph{heuristic} baselines (i.e., simple functions of the VLM classifications or ground-truth annotations which do not require machine learning): (1) the fraction of train set images in a Census tract which are classified positive by the VLM; (2) the number of train set images which are classified positive; (3) whether any train set images are classified positive; (4) whether any train set images are ground-truth annotated positive; and (5) the number of train set images which are ground-truth annotated positive. Second, we compare to \emph{supervised learning} baselines which are trained on the train set to predict the fraction of images which are classified as flooded, and the number of ground-truth annotated flooded images, from the same set of flood-relevant features our Bayesian model uses (Appendix \ref{si:bayesian_modeling}). We fit both linear regression and random forest models. Finally, we compare a \emph{graph smoothing} baseline, which applies Laplacian smoothing using the Census tract adjacency matrix. We fully describe all baselines in Appendix \ref{si:baselines}. 

Our Bayesian model outperforms all baselines on all considered metrics (\autoref{tab:baselines}), demonstrating it provides benefit over alternative ways of aggregating the VLM annotations. 

\begin{table}[htb!]
\small
\begin{tabular}{l>{\raggedright}p{1.6cm}p{1.6cm}p{1.6cm}}
\toprule
Method & Pearson $r$ [frac +\newline classifications] & AUC\newline[any ground-truth +] & AUC\newline[any +\newline classifications]\\
\midrule
Frac. + classifications & $0.39 \pm 0.07$  & $0.76 \pm 0.01$ & $0.67 \pm 0.01$ \\
Any + classfications? & $0.22 \pm 0.02$ & $0.76 \pm 0.01$ & $0.67 \pm 0.01$ \\
\# + classifications & $0.27 \pm 0.03$ & $0.77 \pm 0.01$ & $0.67 \pm 0.01$ \\
Any + annotations? & $0.22 \pm 0.03$ & $0.64 \pm 0.02$ & $0.57 \pm 0.01$ \\
\# + annotations & $0.23 \pm 0.03$ & $0.64 \pm 0.02$ & $0.57 \pm 0.01$ \\
OLS & $0.20 \pm 0.01$ & $0.77 \pm 0.01$ & $0.69 \pm 0.01$ \\
Random Forest & $0.41 \pm 0.07$ & $0.80 \pm 0.02$ & $0.71 \pm 0.01$ \\
Laplacian smoothing & $0.42 \pm 0.07$ & $0.82 \pm 0.01$ & $0.74 \pm 0.01$ \\
\textbf{\ourmethod} & \textbf{0.57} $\pm$ \textbf{0.04} & \textbf{0.88} $\pm$ \textbf{0.01} & \textbf{0.79} $\pm$ \textbf{0.01} \\
\bottomrule
\end{tabular}
\caption{Applying the Bayesian model to the VLM classifications improves out-of-sample prediction of three outcomes relative to baselines. From left to right, the columns report Pearson correlation with fraction of images classified positive; AUC for predicting whether there are any ground-truth positive annotations; and AUC for whether there are any positive VLM classifications. We report the mean and standard deviation across 10 random train/test splits.}
\label{tab:baselines}
\end{table}

\vspace{-12pt}

\subsubsection{Model predictions remain stable even with very few ground-truth annotations}

Because ground-truth human annotations can be expensive to produce in some settings, we investigate whether the flood risk predictions of our Bayesian model, $\floodrisk$, remain stable even with very few ground-truth annotations, showing that the model can be reliably applied even when annotations are sparse. Specifically, we refit the Bayesian model on datasets where the number of ground-truth annotations have been downsampled by a factor of 2$\times$ - 20$\times$; 20$\times$ downsampling corresponds to only 25 annotated positives and 25 annotated negatives. We find that the model's predictions on these downsampled datasets remain highly correlated with predictions on the full dataset (between 0.89 - 0.94 across all downsampling ratios). This suggests that our Bayesian approach can be applied even in settings where very few ground-truth annotations can be collected. Further, because the Bayesian model yields measures of uncertainty on all estimates, it naturally provides principled estimates of the stability of model predictions, guiding the collection of additional annotations if needed. 

\begin{figure*}[h!]
    \centering
    \begin{subfigure}[b]{0.23\textwidth}
        \centering
        \includegraphics[width=\textwidth]{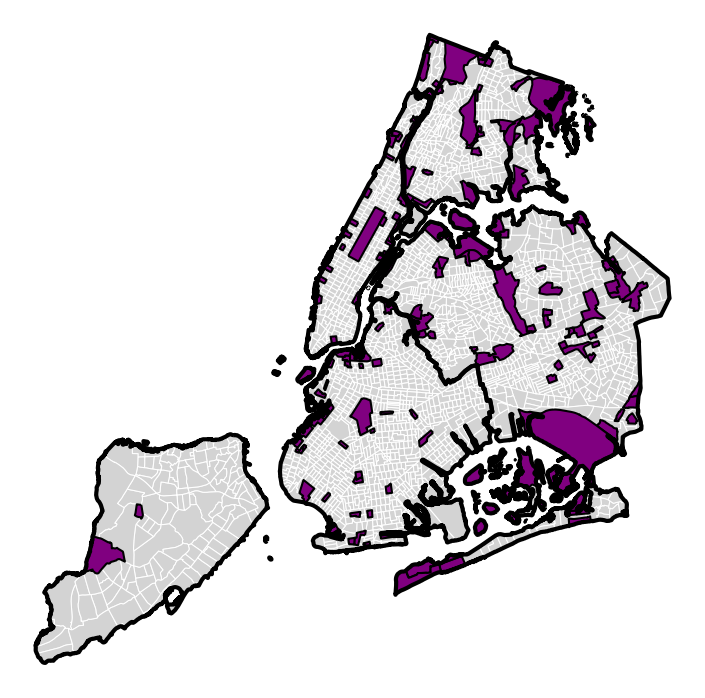}
        \caption{High \ourmethod risk, but\\ no 311 reports.}
        \label{fig:no_311}
    \end{subfigure}
    \begin{subfigure}[b]{0.23\textwidth}
        \centering
        \includegraphics[width=\textwidth]{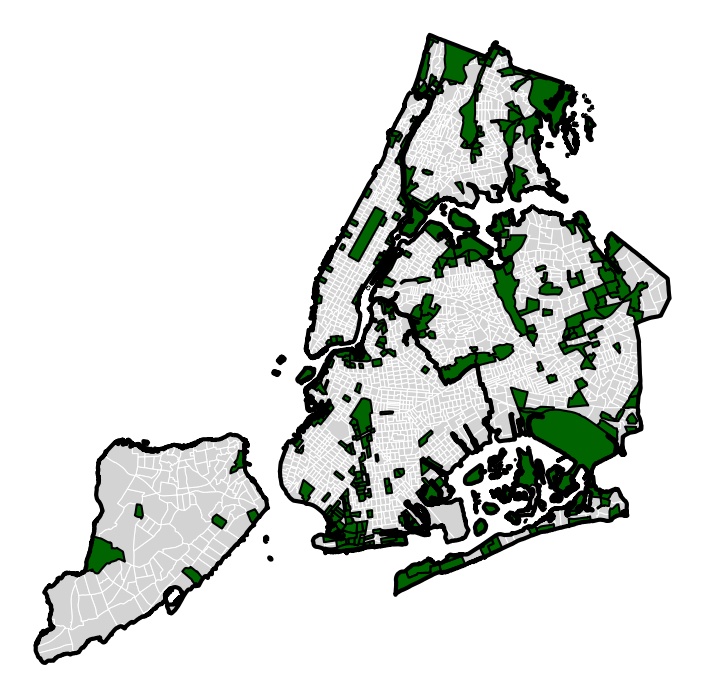}
        \caption{High \ourmethod risk, but\\ no sensors.}
        \label{fig:no_floodnet}
    \end{subfigure}
    \hfill
    \begin{subfigure}[b]{0.23\textwidth}
        \centering
        \includegraphics[width=\textwidth]{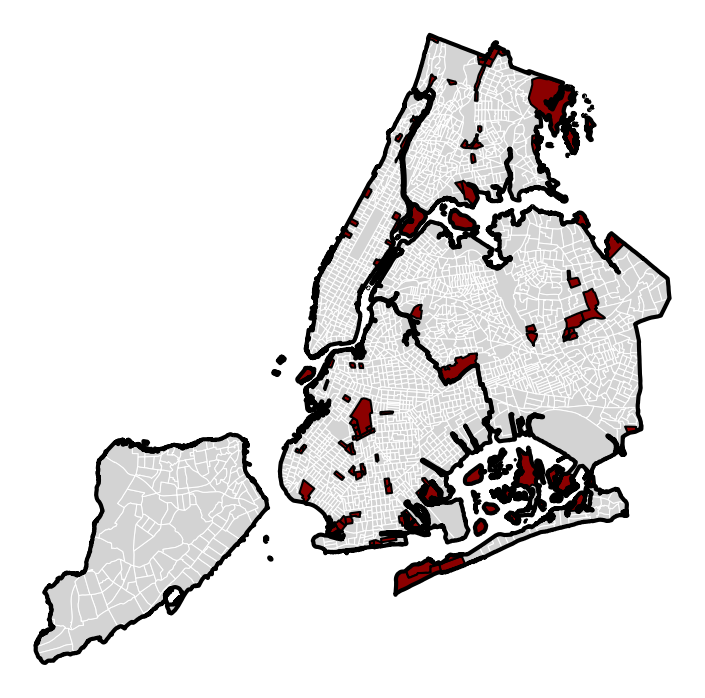}
        \caption{High \ourmethod risk, but\\ no stormwater predictions.}
        \label{fig:no_dep}
    \end{subfigure}
    \hfill
    \begin{subfigure}[b]{0.23\textwidth}
        \centering
        \includegraphics[width=\textwidth]{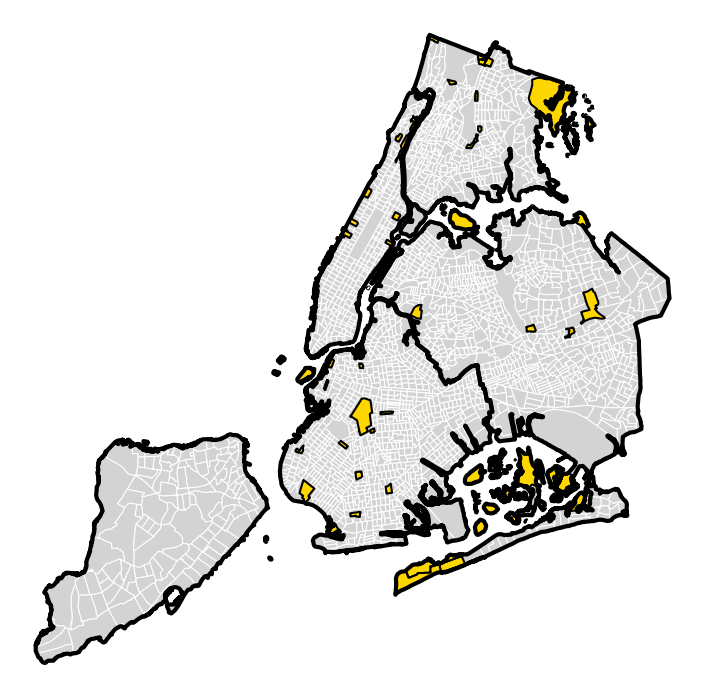}
        \caption{\ourmethod is only flood risk source.}
        \label{fig:isolated_signal}
    \end{subfigure}
    
    \caption{\ourmethod can identify locations at risk for flooding which are missed by three currently used methods. (a) Census tracts with high \ourmethod risk, but no 311 flooding reports; (b) tracts with high \ourmethod risk but no FloodNet sensors; (c) tracts with high \ourmethod risk but no predicted stormwater accumulation; (d) tracts with high \ourmethod risk and no signal from any of the three existing methods.}
    \label{fig:flood_comparison}
\end{figure*}

\subsubsection{Inferred flood risk correlates with known markers of flood risk}
\label{sec:external_validation}
We assess whether our Census-tract-specific estimates of flood risk correlate with the external markers of flood risk discussed in \S \ref{sec:additional_datasets} (Figure \ref{fig:external-covs-vs-risk}). For this analysis, we fit our Bayesian model without incorporating any of these external features, so we are assessing the model's consistency with external flood risk markers it does not have access to. We define a census tract $c$ as ``high \ourmethod risk'' if either $c \in C_\text{confirmed}$, where $C_\text{confirmed}$ is the set of all tracts with a confirmed ground-truth annotated flood image, or if $\floodrisk > t$, where $t$ is the 25th percentile of $r_c$ among all tracts in $C_\text{confirmed}$. We find that \ourmethod's predictions indeed predict external markers of flood risk: its high-risk Census tracts are 1.4$\times$ likelier to have a 311 report and 2.0$\times$ likelier to have a FloodNet sensor. Their minimum elevation is 2.0$\times$ lower and they have 1.3$\times$ larger shallow stormwater accumulation zones as assessed by the Department of Environmental Protection and 1.2$\times$ more deep stormwater accumulation zones. (All differences are statistically significant except deep stormwater accumulation zones (p=0.068); $p<0.005$, t-test).   %

\subsection{Improving flood detection in New York City} 

Having validated \ourmethod, we show it can be applied to three important use cases: detecting flooded areas missed by existing methods; quantifying biases in 311 reports; and suggesting new locations for flood sensors. These applications are informed by our conversations with government decision-makers as well as with academic-government partnerships like FloodNet. 
\label{sec:improving_flood_detection}

\subsubsection{Detecting flooded areas missed by existing methods}
Our model can identify Census tracts at risk for flooding which are missed by methods currently used by urban decision-makers (\autoref{fig:flood_comparison}). We quantify the number of Census tracts that are predicted high-risk by \ourmethod but do not have a flood-related 311 report, a FloodNet sensor, or predicted stormwater accumulation. (For this analysis, we define tracts with high \ourmethod risk as in \S\ref{sec:external_validation}.)%

1,003,940 people live in the Census tracts with high \ourmethod risk, comprising 12\% of New York City's population. Of these, \numNonThreeOneOneCovered\space people live in Census tracts with no flooding-related 311 reports; \numNonFloodNetCovered\space people live in Census tracts with no FloodNet sensors; \numNonDEPCovered\space people live in Census tracts with no predicted stormwater accumulation; and \numNewResidentsCovered\space people live in Census tracts with no indicator of flood risk from any of these methods.\footnote{Supplementary Figure \ref{fig:confirmed-flooding-map-comparison} reports a version of this analysis redefining ``high flood risk'' tracts as those at least one ground-truth confirmed flooded image ($y=1$); this similarly identifies many tracts which are missed by current flood detection methods, though not as many as those identified by our Bayesian model, highlighting the benefits of our approach.} Collectively, these results indicate that our model can identify large populations of people who face flood risks currently overlooked by some or all of the existing flood detection methods. 

\subsubsection{Quantifying biases in 311 reports}

Previous work has raised concerns that 311 reports may display demographic biases, with some neighborhoods less likely to report incidents when they occur \cite{agostini_bayesian_2024, kontokosta_bias_2021, kontokosta_equity_2017}. We investigate whether our model can quantify these biases. Specifically, we conduct a \emph{risk-adjusted logistic regression}~\cite{jung2018mitigating}, which assesses whether there are demographic disparities in 311 reporting patterns across Census tracts which cannot be explained by our model's estimated flood risk:
\begin{align*}
p(\text{Tract $c$ has 311 report}) = \text{logit}^{-1}(\gamma + \beta_r \floodrisk + \beta_d d_c)
\end{align*}

\begin{figure}[h!]
    \includegraphics[width=0.45\textwidth]{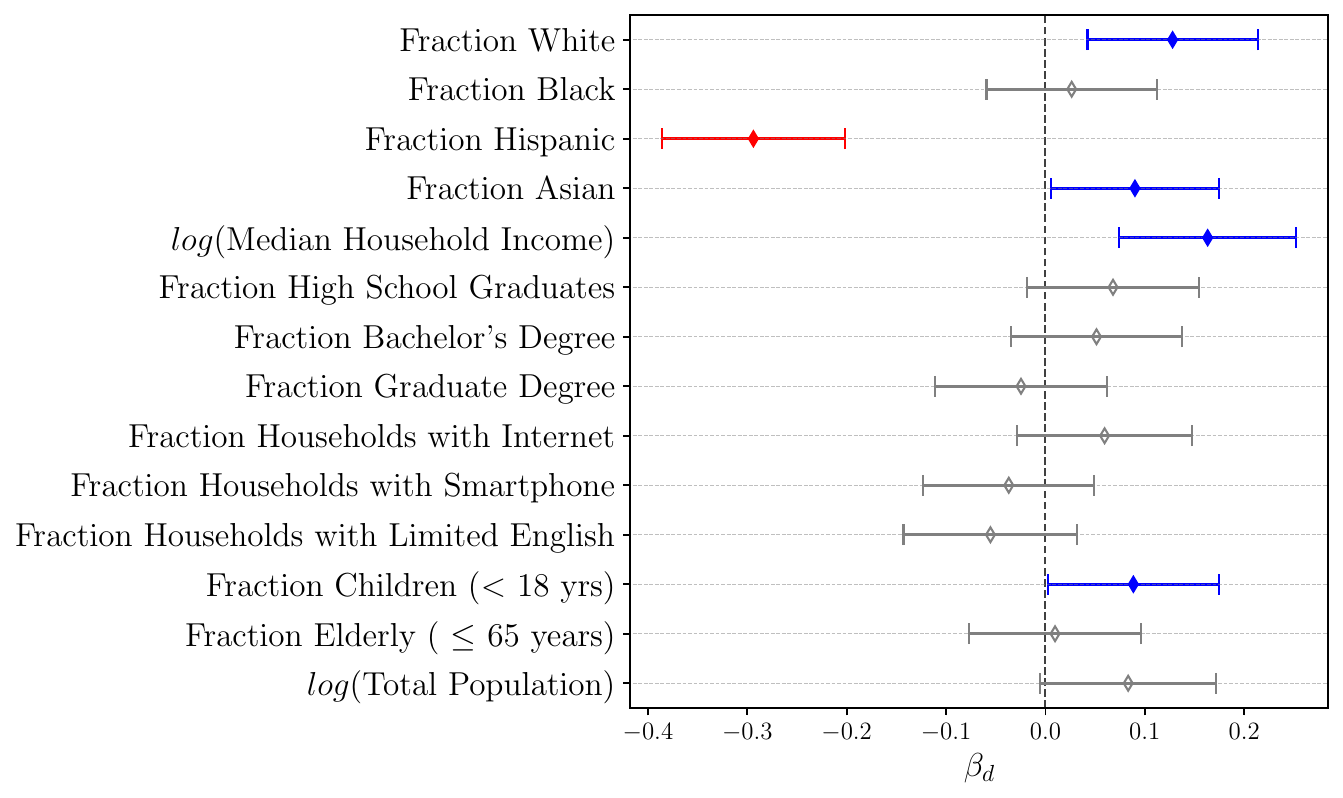}
    \caption{Demographic coefficients for the risk-adjusted regression reveal biases in 311 reporting patterns. 95\% confidence intervals are plotted; all demographic features are z-scored, so coefficients are in units of standard deviations of each feature.}
    \label{fig:311-demo-biases}
\end{figure}

where $\gamma$ is an intercept term; $\floodrisk$ is our model's estimate of flood risk in tract $c$; $d_c$ is a demographic feature\footnote{All demographic data comes from the American Community Survey 2023 5-Year Estimates.} (e.g., the fraction of the tract which is white); and the $\beta$s are the regression coefficients. \autoref{fig:311-demo-biases} plots the estimated demographic coefficients $\beta_d$. Controlling for flood risk, we find that Census tracts with larger fractions of white and Asian residents, lower fractions of Hispanic residents, higher average household incomes, and higher fractions of children are statistically significantly likelier to have a 311 report. These findings accord with past work providing evidence of biases in 311 reporting patterns, and show that our model can be usefully applied to audit existing methods of flood detection.\footnote{Supplementary Figure \ref{fig:bias-confirmed-flooding-only} repeats this analysis controlling for an alternate measure of flood risk: whether a tract has at least one ground-truth confirmed flooded image ($y=1$); results are similar.}

\subsubsection{Suggesting new locations for sensor placement}

Based on our finding that many Census tracts have no flood sensors, but high predicted flood risk, we provide a proof-of-concept illustration that our model can be applied to identify tracts which might benefit from the placement of a new sensor.

We assume that if a sensor is placed in a given tract, it can detect a flood in all tracts within a $k$-hop neighborhood, because floods are spatially correlated. We set $k=1$ in our experiments, but our framework can easily be applied to other $k$ (as well as to incorporate additional considerations like the population of a tract, equity in sensor placement, etc). Given the current locations of $T$ sensors in a set of Census tracts $\mathcal{S}^t \triangleq \{c_1, c_2, ..., c_T\}$, our task is to place an additional $U$ sensors in a set of Census tracts $\mathcal{S}^u \triangleq \{c_{T + 1}, c_{T + 2}, ..., c_{T + U}\}$ to maximize the sum of flood risk $\floodrisk$ in covered areas: 
\begin{align*}
\sum_{c \in A(\mathcal{S}^{t} \cup \mathcal{S}^u)} \floodrisk
\end{align*}

\begin{figure}[h!]
    \includegraphics[width=0.4\textwidth]{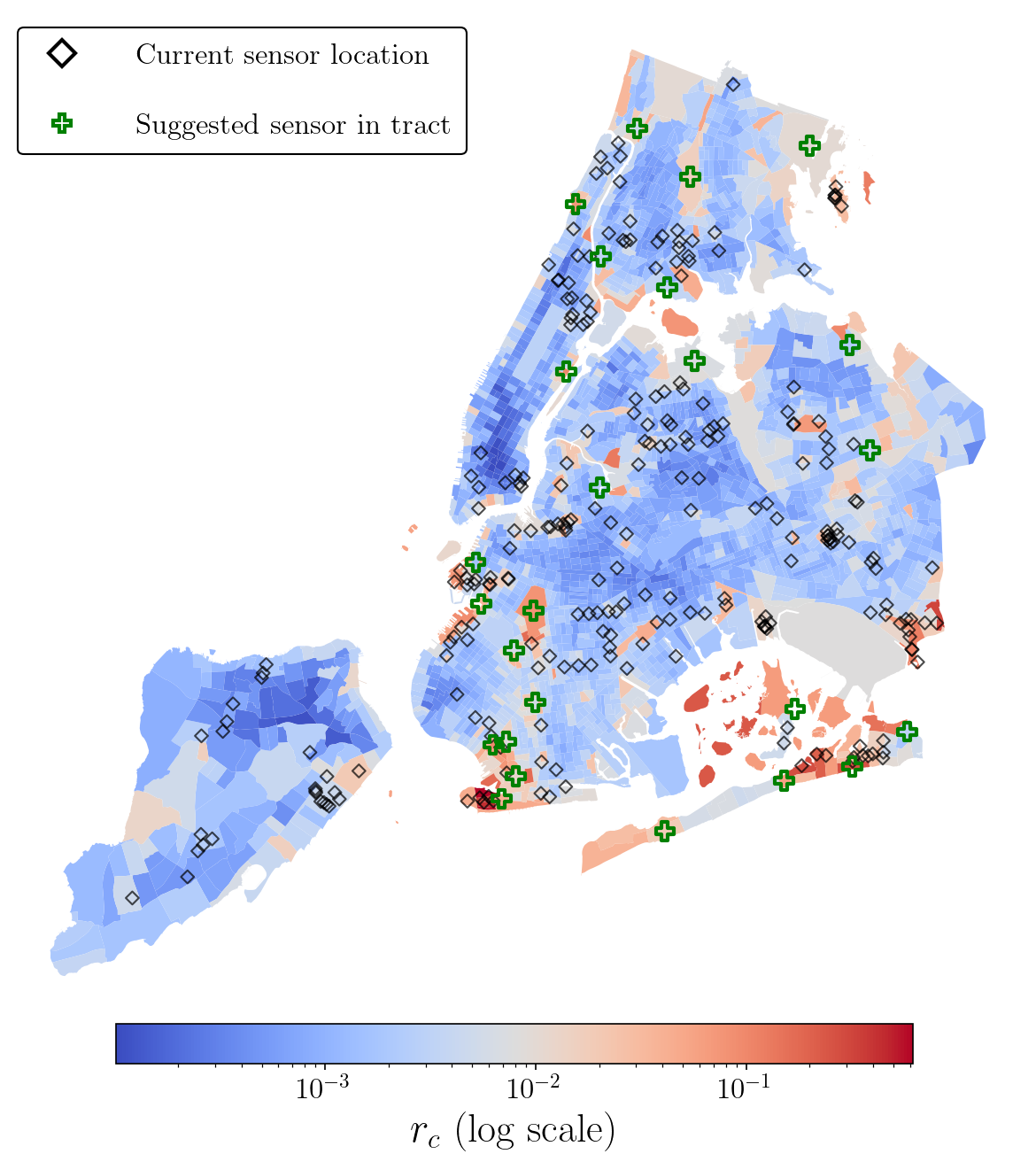}
    \caption{Existing FloodNet sensor locations (black diamonds), and suggested locations for new sensors (green crosshairs).}
    \label{fig:floodnet-placement}
\end{figure}

where $A(\mathcal{S})$ denotes all Census tracts in the neighborhood of set $\mathcal{S}$. This is a \emph{weighted maximum coverage} problem~\cite{nemhauser1978analysis} in which we are given a collection of sets, and our goal is to choose $U$ sets such that the weighted sum of elements is maximized. Here, each set is the tracts covered by a sensor in a given location, and the weight for each tract is $\floodrisk$. This problem is NP-hard, but because the optimization objective is submodular, the greedy solution achieves an approximation ratio of $1 - \frac{1}{e}$, and is often used. At each iteration, we greedily choose the Census tract that maximizes the sum of $\floodrisk$ in newly covered tracts. We plot the locations chosen by this procedure in Figure \ref{fig:floodnet-placement}, setting $U=25$. We are submitting our suggested locations to the FloodNet collaboration \cite{mydlarz_floodnet_2024} as part of our ongoing conversations with them regarding sensor placements. They expressed interest in our data and methodology as one valuable source of signal to supplement their ongoing placement methodology, which is largely driven by community engagement, stakeholder needs, and equity considerations \cite{ceferino_developing_2023}. %

\section{Discussion}

In this work, we developed a novel two-stage method, \ourmethod, that combines modern VLMs with classical Bayesian spatial modeling to detect urban incidents such as street flooding.  In the first stage, we conduct zero-shot classification using a pre-trained VLM to identify flooding in street images, avoiding the need for large labeled datasets. In the second stage, the results from this classification are integrated into a Bayesian spatial model; this provides the benefits of classical statistical methods, including principled measures of uncertainty, spatial smoothing, and incorporation of external datasets. We show that our approach can effectively detect floods and improves on baseline approaches. We apply our methodology to detect floods missed by existing urban detection methods; reveal biases in current approaches; and suggest locations for new flood sensors.

There are several natural directions for future work. Within urban data science, one might expand our approach to additional cities and flood events. Creating a model which could run in real time, providing insight into ongoing floods, might also offer significant benefits to decision-makers. One might also expand our approach to detect other types of urban incidents, like unpermitted sidewalk scaffolding~\cite{shapira_fingerprinting_2024}, double parking, or out-of-place garbage; a significant benefit of our methodology is that it relies only on zero-shot detection, avoiding the need for large labeled datasets and easing its application to new types of incidents. Methodologically, there are also avenues for future work, including experimenting with alternate VLMs or model prompts and using temporal or hierarchical Bayesian models which allow for change over time and incorporation of additional storms. More broadly, our results showcase how Bayesian modeling of zero-shot foundation model annotations represents a promising paradigm which combines the power of foundation models with the benefits of classical statistical methods. This paradigm has broad potential applicability in the many settings in the natural and social sciences where foundation models are increasingly being used for annotation.

\paragraph{Code release.} All code and aggregated data for replicating our analysis (including our VLM inferences) are available at this \textbf{\href{https://github.com/mattwfranchi/street-flooding}{ GitHub repository}}\color{black}.

\begin{acks}
 We thank Gabriel Agostini, Sidhika Balachandar, Serina Chang, Zhi Liu, and Anna McClendon for useful discussion and feedback. We thank Nexar for data access under research evaluation and project support. We thank Anthony Townsend and Michael Samuelian for project support. We thank the NYC Department of Environmental Protection for helpful discussions. We thank the FloodNet team for helpful discussions and access to FloodNet data. We thank the Digital Life Initiative, the Urban Tech Hub at Cornell Tech, a Google Research Scholar award, an AI2050 Early Career Fellowship, NSF CAREER \#2142419, NSF CAREER IIS-2339427, a CIFAR Azrieli Global scholarship, a gift to the LinkedIn-Cornell Bowers CIS Strategic Partnership, the Survival and Flourishing Fund, and the Abby Joseph Cohen Faculty Fund for funding. 

\end{acks}

\clearpage

\bibliographystyle{ACM-Reference-Format}
\bibliography{sample-base,matt,emma,census}

\clearpage

\appendix
\section{Details of empirical setting}

\setcounter{figure}{0}
\renewcommand{\figurename}{Figure}
\renewcommand{\thefigure}{S\arabic{figure}}
\setcounter{table}{0}
\renewcommand{\tablename}{Table}
\renewcommand{\thetable}{S\arabic{table}}

\subsection{Primary analysis dataset}
We provide additional details on the September 29, 2023 flooding event in New York City on which we conduct our primary analysis. This flooding event was triggered by an intense storm system that dumped several inches of rain in just a few hours, with some areas receiving over 7 inches \cite{the_associated_press_new_2023}. The Governor of New York State declared a state of emergency for New York City and adjacent areas.
The flooding severely disrupted the city's transportation infrastructure: service was suspended on multiple subway lines as water poured into stations and tunnels \cite{vivian_camacho_flood_2023}; major highways experienced significant flooding, stranding motorists; airport terminals were closed due to flooding~\cite{staff___laguardia_2023}.
The intense rainfall overwhelmed the city's drainage system, which wasn't designed to handle such extreme precipitation events. Many neighborhoods experienced flash flooding, with water entering homes and businesses. Videos shared on social media showed cars partially submerged on major streets and people wading through waist-deep water; 28 individuals had to be rescued \cite{news_28_nodate}.

Climate change is increasing the frequency and intensity of such extreme weather events in urban areas \cite{brody_rising_2007, newman_global_2023}. This flooding event that occurred just two years after Hurricane Ida caused devastating flooding in New York City, raising questions about the city's infrastructure resilience and adaptation strategies \cite{the_associated_press_new_2023}.
The flooding also disproportionately impacted some of the city's most vulnerable areas where drainage infrastructure is older or inadequate. Many basement apartments, often occupied by lower-income residents, were flooded, echoing similar patterns seen during Hurricane Ida.
The event led to renewed calls for infrastructure improvements and better stormwater management systems, as well as discussions about how to better protect vulnerable communities from extreme weather events. The increasing severity and prevalence of such flooding events, and recognition of the need for improved detection methods, motivates our analysis in this paper, informed by our conversations with city decision-makers.

\begin{table*}
\small
\begin{tabular}{p{0.5\linewidth}p{0.5\linewidth}}
\toprule
\textbf{Description} & \textbf{Data Source} \\
\midrule
\multicolumn{2}{l}{\textit{Census and ACS Characteristics}} \\
\midrule
Total Population in Census tract & ACS DP05: Demographic and Housing Estimates \cite{Census2023ACSDP5Y2023.DP05} \\
Non-Hispanic White Population & ACS DP05: Demographic and Housing Estimates \cite{Census2023ACSDP5Y2023.DP05} \\
Non-Hispanic Black Population & ACS DP05: Demographic and Housing Estimates \cite{Census2023ACSDP5Y2023.DP05} \\
Hispanic Population & ACS DP05: Demographic and Housing Estimates \cite{Census2023ACSDP5Y2023.DP05} \\
Non-Hispanic Asian Population & ACS DP05: Demographic and Housing Estimates \cite{Census2023ACSDP5Y2023.DP05} \\
Number of Households with Internet Access & ACS 2801: Types of Computers and Internet Subscriptions \cite{Census2023ACSST5Y2023.S2801} \\
Number of Households with Smartphone Access & ACS 2801: Types of Computers and Internet Subscriptions \cite{Census2023ACSST5Y2023.S2801} \\
Median Annual Household Income (USD) & ACS S1901: Income in the Past 12 Months \cite{Census2023ACSST5Y2023.S2801} \\
Number of High School Graduates & ACS S1501: Educational Attainment \cite{Census2023ACSST5Y2023.S1501} \\
Number of Bachelor's Degree Holders & ACS S1501: Educational Attainment \cite{Census2023ACSST5Y2023.S1501} \\
Number of Graduate Degree Holders & ACS S1501: Educational Attainment \cite{Census2023ACSST5Y2023.S1501} \\
Number of Limited English Proficiency Households & ACS 1602: Limited English Speaking Households \cite{Census2023ACSST5Y2023.S1602} \\
\midrule
\multicolumn{2}{l}{\textit{Physical Geography}} \\
\midrule
Minimum Elevation in Census tract (feet) & 1 foot NYC Digital Elevation Model (DEM) \cite{nyc_opendata_1_2024} \\
Maximum Elevation in Census tract (feet) & 1 foot NYC Digital Elevation Model (DEM) \cite{nyc_opendata_1_2024} \\
Mean Elevation in Census tract (feet) & 1 foot NYC Digital Elevation Model (DEM) \cite{nyc_opendata_1_2024} \\
Geographic Area of Census tract (square feet) & 2020 Census tracts Shapefile, NYC DCP \cite{nyc_dcp_census_2025} \\
\midrule
\multicolumn{2}{l}{\textit{Flood Infrastructure and Risk}} \\
\midrule
Number of Flooding-Related 311 Complaints & 311 Service Requests, 2010-Present \cite{nyc_opendata_311_2025} \\ 
Number of FloodNet Sensors Installed & FloodNet Team (NYU, CUNY) \\
Area of Shallow Flooding (4in--1ft) Under Moderate Rain (2.13in/hr) & NYC DEP Stormwater Maps \cite{nyc_department_of_environmental_protection_2024_nodate} \\
Area of Deep Flooding (>1ft) Under Moderate Rain (2.13in/hr) & NYC DEP Stormwater Maps \cite{nyc_department_of_environmental_protection_2024_nodate} \\
Fraction of Total Area with Shallow Flooding & NYC DEP Stormwater Maps \cite{nyc_department_of_environmental_protection_2024_nodate} \\
Fraction of Total Area with Deep Flooding & NYC DEP Stormwater Maps \cite{nyc_department_of_environmental_protection_2024_nodate} \\
\bottomrule
\end{tabular}
\caption{Fields from external datasets used in our analysis.}
\label{tab:data-dictionary}
\end{table*}

\subsection{Additional data processing details.}
\label{si:geospatial}

\begin{figure}
    \includegraphics[width=0.48\textwidth]{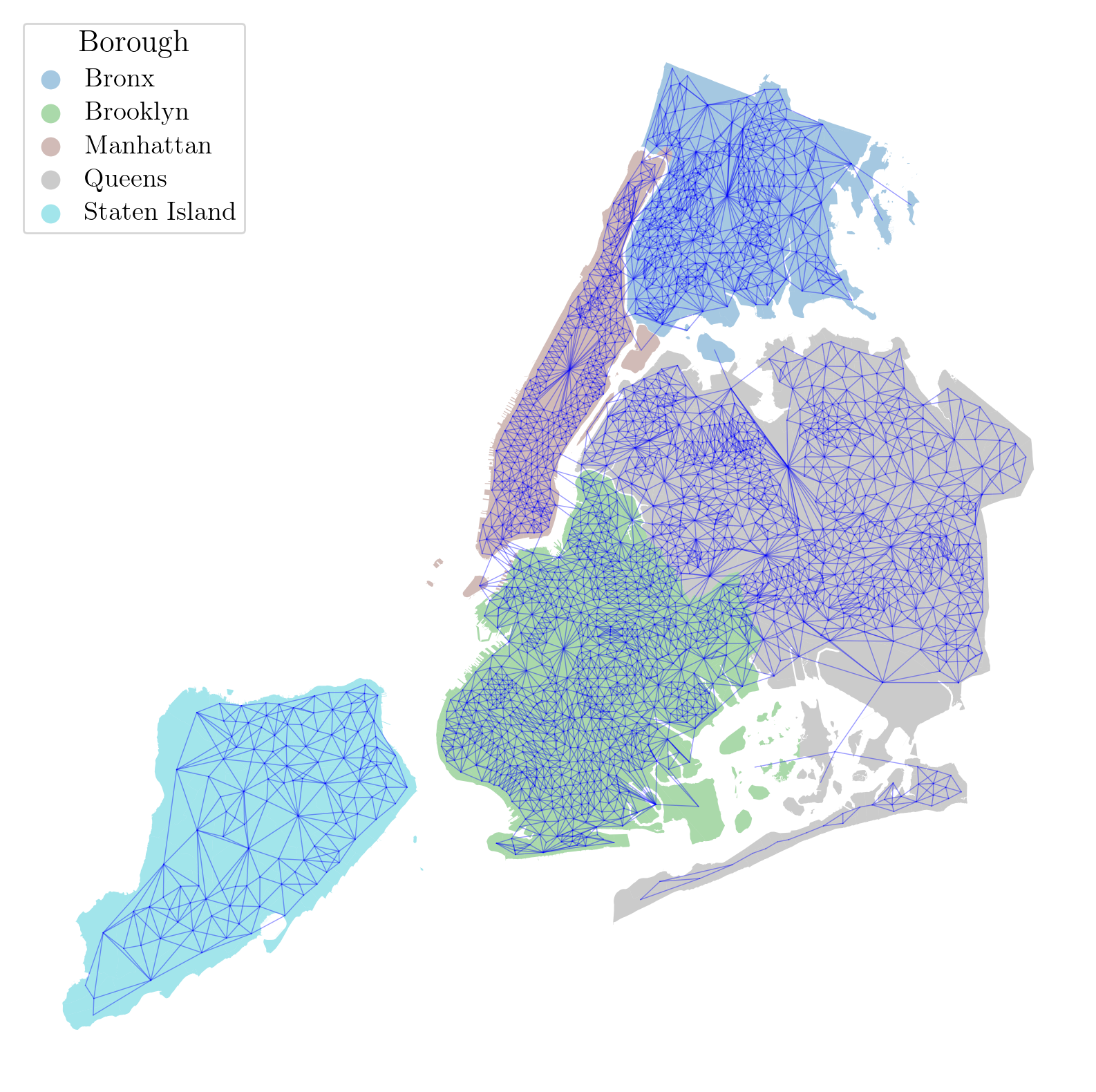}
    \caption{Our adjacency network of NYC's Census tracts.}
    \label{fig:adj}
\end{figure}
\paragraph{Census tract adjacency matrix.} We describe our approach in assessing adjacent Census tracts in New York City, as the city's topography and dense development have produced intricate Census geographies. Further, with the flooding-oriented nature of our work, the accuracy of spatial adjacency relationships is important. 

We use the water-clipped version of the 2020 NYC Census tracts, provided by the NYC Department of City Planning \cite{nyc_dcp_census_2025}. Using tracts with water areas included creates inaccurate adjacency relationships, such as tracts in Downtown Manhattan being neighbors to Governor's Island. 
We generate neighbor relationships through geometric processing; we buffer (expand) each tract by 500 feet, and then assign adjacency between tracts that intersect. We visualize our adjacency matrix in \autoref{fig:adj}.

\begin{figure}[h!]
    \centering
    \includegraphics[width=0.25\textwidth]{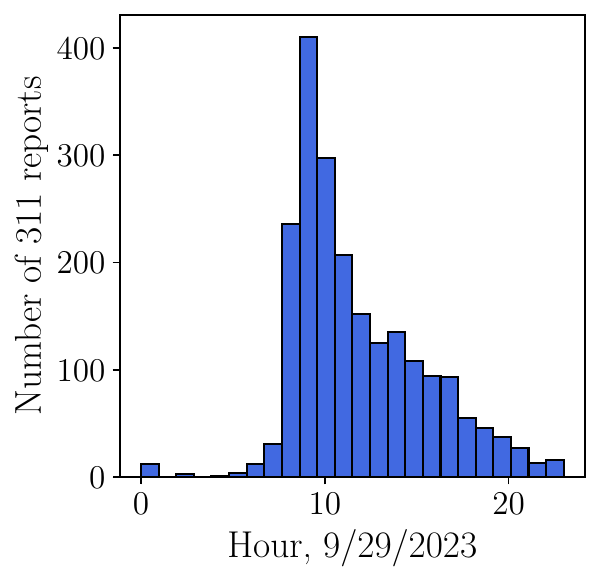}
    \caption{By-hour frequency histogram of flooding-related 311 reports on September 29, 2023 in NYC.}
    \label{fig:311-hist}
\end{figure}

\label{si:datasets}
\paragraph{311 reports.} We denote the following complaint types as flooding-related for the purposes of our analysis: sewer backup, street flooding, catch basin clogged/flooded, manhole overflow, and highway flooding. We include reports from the entire day of September 29, 2023. The frequency of flooding-related 311 reports (see \autoref{fig:311-hist}) corresponds with the progress of the storm event \cite{nyc_comptroller_is_2024} .

\paragraph{Elevation data.} We downsample the 1-foot Digital Elevation Model of NYC by a factor of 10 because it accelerates performance without significantly compromising accuracy at the granularity of a Census tract. 

\section{Additional VLM details}
\label{si:additional_vlm_details}

\subsection{Measurement of VLM performance}

For all VLMs, including baseline models, we measure performance on our primary dataset by manually annotating a random subset of classified positives and a random subset of classified negatives. For our preferred VLM, we additionally assess performance on three additional days (as described in \S\ref{sec:dashcam_data}) by manually annotating 250 randomly-sampled classified negatives and 250 randomly-sampled classified positives (or, in cases where there are fewer than 250 classified positives, all classified positives) -- see \autoref{tab:other-days-performance}.

We only annotate an image as flooded if it unambiguously shows flooding. In very rare cases, technical artifacts render this ambiguous, including cases where (a) the view of the forthcoming street is obscured by the vehicle dashboard or (b) the dashcam is mispositioned, producing images that do not depict any part of the street; we mark these as negative. We similarly annotate images with visually ambiguous flooding (e.g., ambiguous reflections from sunlight) as negative. 

\begin{table}[h!]
\small
\begin{tabular}{lcccc}
\toprule
 & 9/29/23 & 12/18/23 & 1/10/24 & 2/10/24 \\
& New York & New York & New York & San Francisco \\
\midrule
$p(y=1|\hat y = 1)$ & 0.658 & 0.702 & 0.812 & 0.143 \\
$p(y=1|\hat y = 0)$ & 0.006 & 0.000 & 0.000 & 0.000 \\
\bottomrule
\end{tabular}
\caption{Validation of VLM performance across multiple days and locations. Results reported are for our preferred model (Cambrian-1-13B). Classified positives ($\hat y = 1$) are much likelier to show flooding ($y = 1$) than classified negatives across all four days.}
\label{tab:other-days-performance}
\end{table}

\vspace{-1cm}

\subsection{VLM baselines}
\label{si:engineering}

We compare the performance of our preferred VLM (Cambrian-13B) to several alternate VLM architectures with zero-shot prompting: CLIP; DeepSeek's Janus Pro VLM; and Cambrian-1-8B. For all baselines, we experiment with multiple prompts and report the highest-performing configuration for each VLM in \autoref{tab:vlm-baselines}. Our preferred model achieves superior performance to the baselines. We estimate performance of each baseline by sampling a random subset of 250 classified positives, and 250 classified negatives, and obtaining ground-truth manual annotations. 

We also compared our zero-shot prompting method to a supervised learning approach using noisy labels. Specifically, we first used CLIP (zero-shot) to identify candidate flooded images; then used GPT-4V (zero-shot) to further filter down the set, producing a set of noisy positives; and then fine-tuned a ResNet to distinguish between the noisy positives and all other images. (We did not use GPT-4V to annotate all images because it would impose a prohibitive cost on a dataset of our size.) We found that this method achieved inferior performance to our preferred approach (\autoref{tab:vlm-baselines}) at the cost of considerable additional complexity, and thus did not pursue it further.

\subsection{Alternate VLM prompts}
For each VLM we test, we assess multiple prompts and report results from the one which yields optimal performance. For the VLM we used for our primary analysis, Cambrian-13B, we compared performance of two prompts: (a) \textit{Does the street in this image show more than a foot of standing water?} and (b) \textit{Does this image show a flooded street?} We found that the latter prompt classified many more images as positive, resulting in slightly higher recall but much lower precision, and thus used the former.

\section{Additional Bayesian modeling details} 
\label{si:bayesian_modeling}

\subsection{Model priors}

We place the following weakly informative priors on model parameters: 
\begin{align*}
    \alpha &\sim \text{Normal}(-5, 2)\\
    \beta &\sim \text{Normal}(0, 2)\\
    \theta_{\hat y = 1 | y = 1} &\sim \text{Logit-Normal}(0, 2)\\
    \theta_{\hat y = 1 | y = 0} &\sim \text{Logit-Normal}(0, 2)
\end{align*}
The negative-centered prior on $\alpha$ reflects the prior belief that most locations are not flooded. We place an ordered vector constraint on $\theta$ to enforce the assumption that $\theta_{\hat y = 1 | y = 1} > \theta_{\hat y = 1 | y = 0}$. 

Our model of $p(y)$ also includes an ICAR spatial component $\phi$: 
\begin{equation*}
p(y=1|C=c)=\text{logit}^{-1}(\alpha + X_c\beta + \phi \cdot \sigma_\phi)
\end{equation*}
ICAR models are commonly used to capture spatially correlated data. We implement the ICAR component, as is standard, by incrementing the log probability using the the pairwise difference formula~\cite{morris2019bayesian}: 
\begin{equation*}\sum_{i, j: A_{ij} = 1, i < j}-\frac{1}{2}(\phi_i - \phi_j)^2
\end{equation*}
where $A$ denotes the adjacency matrix for Census areas. We place a soft sum-to-zero constraint on $\phi$. We place a $\text{Normal}_+(0, 1)$ prior on the standard deviation of the ICAR component, $\sigma_\phi$.

\subsection{Flooding features used in Bayesian model}
\label{sec:flooding_features_in_bayesian_model}

As an input to our Bayesian model (i.e., $X_c$ in the notation above), we use 6 flood features from the external datasets described in \S\ref{sec:additional_datasets}: the number of 311 reports in a Census tract, the number of FloodNet sensors in a tract, the minimum elevation of the tract, the mean elevation of the tract, and the fraction of the tract with shallow and deep flooding in New York City Department of Environmental Protection Stormwater maps. We log-transform all right-skewed features, and z-score all features, prior to using them as inputs to the model, as is standard \cite{friedman_elements_2009}. 

\subsection{Comparison of Bayesian model to baselines} 
\label{si:baselines}

As described in \S\ref{sec:results_validation}, we show that our Bayesian approach improves predictions of where flooded images will occur on a held-out test set, relative to both simple heuristics (e.g., the fraction of images which are classified as positive by the VLM) and machine learning baselines. We describe these baselines below. 

\subsubsection{Heuristic baselines}

We compare to five heuristic baselines: 

\begin{itemize} 
\item \textbf{Fraction of positive classifications}: i.e., the fraction of images in a Census tract which the VLM classifies as positive. 
\item \textbf{Any positive classifications}: 1 if the VLM classifies any images in a Census tract as positive, 0 otherwise. 
\item \textbf{Number of positive classifications}: the number of images in a tract the VLM classifies as positive. 
\item \textbf{Any positive ground-truth annotations}: 1 if there are any images in a Census tract with ground-truth annotations as positive, 0 otherwise. 
\item \textbf{Number of positive annotations}:  the number of images in a tract with ground-truth positive annotations. 
\end{itemize} 
We do not use the fraction of positive ground-truth annotations as a baseline because most tracts have no ground-truth annotations at all and for them this baseline is not well-defined.

\subsubsection{Supervised learning baselines}
For the supervised learning baselines, we use the same set of flood-relevant features which are inputs to our Bayesian model (\S \ref{sec:flooding_features_in_bayesian_model}) to predict (a) the fraction of positive classifications and (b) the number of positive ground-truth annotations. We treat the variable to be predicted as a hyperparameter and report the setting which yields the best performance. We report results from both a linear regression model and a random forest model. 

\subsubsection{Graph smoothing baselines}

We compare to \emph{graph Laplacian smoothing baselines}, which use the graph Laplacian $L \triangleq D - A$ (where $D$ is the diagonal degree matrix, and $A$ is the adjacency matrix) to iteratively smooth a graph \cite{li_deeper_2018,chen_deep_2019}. One iteration of the algorithm updates the value $\textbf{x}$ at each node (in our case, a Census tract) using the following diffusion update: $\textbf{x}_{\text{new}} = \textbf{x}_{\text{old}} - \alpha L \textbf{x}_{\text{old}}$, where $\alpha$ is a step size parameter. The hyperparameters are $\alpha$, the number of smoothing iterations, and the initial value to be smoothed (fraction of positive classifications or number of positive ground-truth annotations); we report the hyperparameter configuration which maximizes performance.

\begin{figure*}[h!]
    \centering
    \begin{subfigure}[b]{0.24\textwidth}
        \centering
        \includegraphics[width=\textwidth]{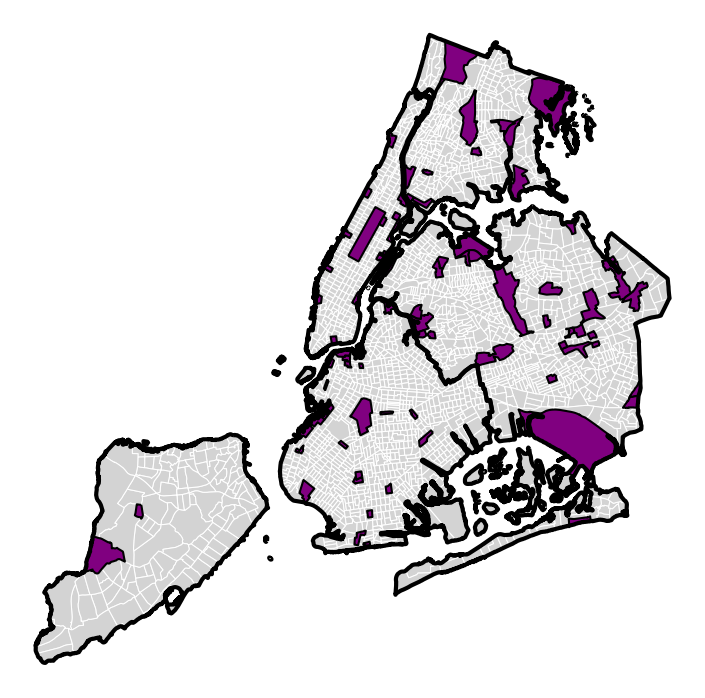}
        \caption{Ground-truth flooded image, no 311 reports}
        \label{fig:confirmed_no_311}
    \end{subfigure}
    \begin{subfigure}[b]{0.24\textwidth}
        \centering
        \includegraphics[width=\textwidth]{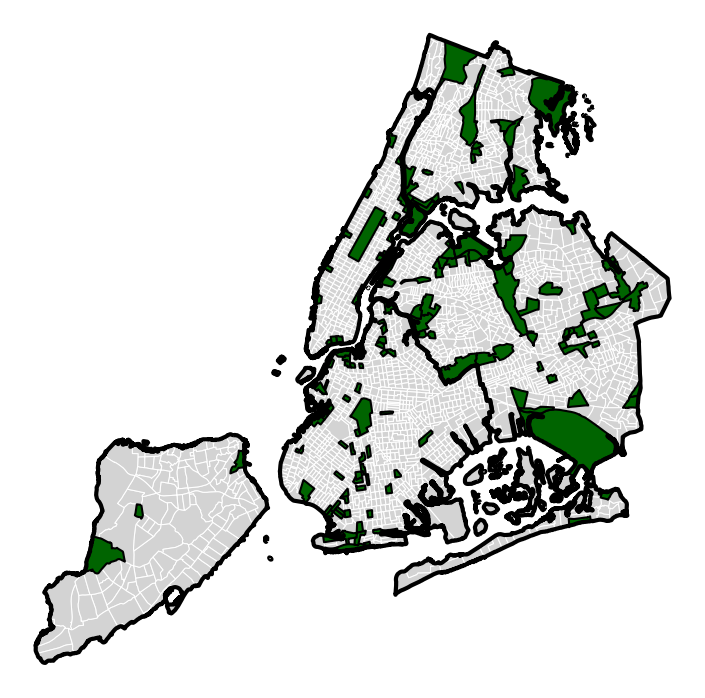}
        \caption{Ground-truth flooded image, no sensors}
        \label{fig:confirmed_no_floodnet}
    \end{subfigure}
    \hfill
    \begin{subfigure}[b]{0.24\textwidth}
        \centering
        \includegraphics[width=\textwidth]{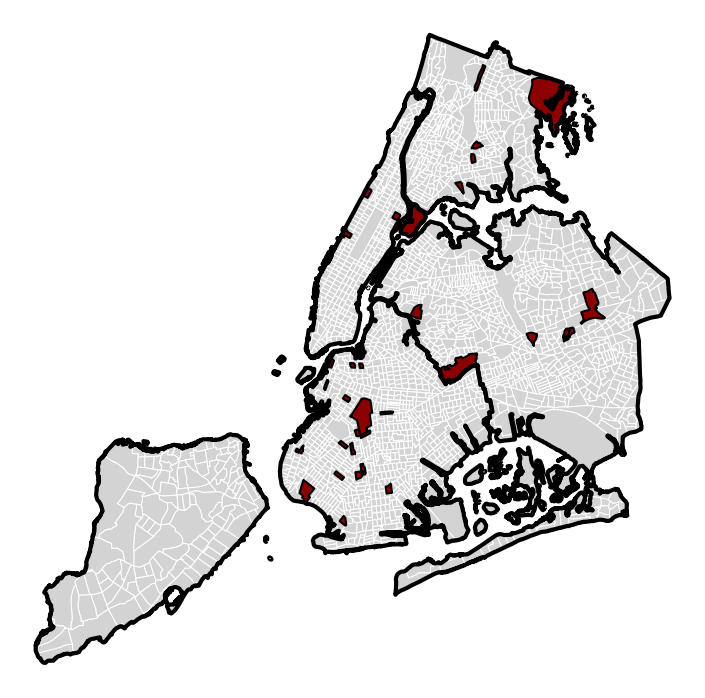}
        \caption{Ground-truth flooded image, no stormwater pred.}
        \label{fig:confirmed_no_dep}
    \end{subfigure}
    \hfill
    \begin{subfigure}[b]{0.24\textwidth}
        \centering
        \includegraphics[width=\textwidth]{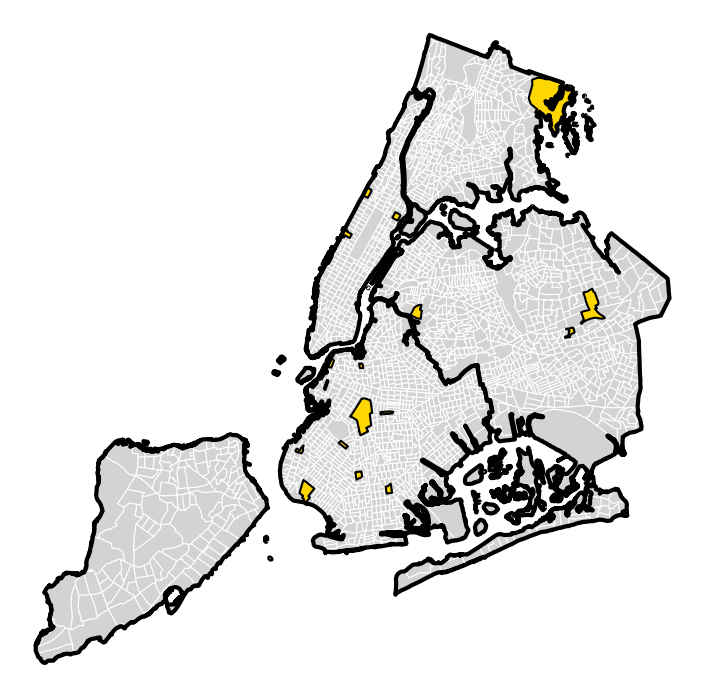}
        \caption{Ground-truth flooded image, no other flooding signal.}
        \label{fig:confirmed_isolated_signal}
    \end{subfigure}
    
    \caption{We repeat the analysis in Figure \ref{fig:flood_comparison}, but defining high-flood risk tracts as only those with a ground-truth annotated flooded image ($y=1$). (a) Ground-truth flooded image, but no 311 flooding reports; (b) ground-truth flooded image, but no FloodNet sensors; (c) ground-truth flooded image, but no predicted stormwater accumulation; (d) ground-truth flooded image, and no signal from any of the three existing methods. 45,229 residents are identified in the final map, relative to \numNewResidentsCovered\space residents when incorporating the high-flood-risk predictions of our Bayesian model, demonstrating the benefits of the Bayesian approach.}
    \label{fig:confirmed-flooding-map-comparison}
\end{figure*}

\subsubsection{Comparison of baseline and Bayesian model performance} 

\begin{figure}[h!]
    \includegraphics[width=0.4\textwidth]{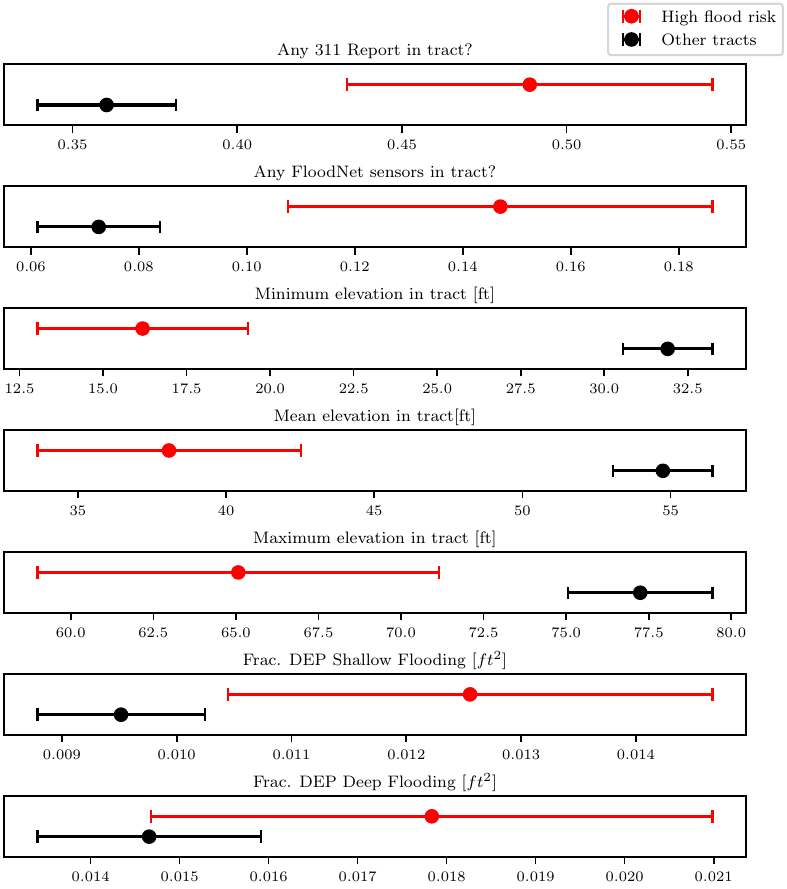}
    \caption{Validation of \ourmethod predictions against external flood-related features. ``High flood risk'' refers to tracts $c$ with either (1) $c \in C_\text{confirmed}$, where $C_\text{confirmed}$ is the set of all tracts with a confirmed ground-truth annotated flood image, or (2) $\floodrisk > t$, where $t$ is the 25th percentile of $r_c$ among all tracts in $C_\text{confirmed}$. For this analysis only, we fit \ourmethod without using any external flood-related features to allow validation of its predictions. All p-values for differences $< 0.005$ except for Frac. DEP Deep Flooding (p=0.068),  t-test.}
    \label{fig:external-covs-vs-risk}
\end{figure}

As described in \S\ref{sec:results_validation}, we compare the performance of our Bayesian model to that of baselines by partitioning the classified images into a train set (which we use to fit the Bayesian model and the baselines on the classifications) and a test set (which we use to assess out-of-sample performance). The inputs to the Bayesian model and the baselines are the number of images in each Census tract with a given ground-truth annotation (positive, negative, or unknown) and a given VLM classification (positive or negative), as well as the flood-risk features $X_c$; some baselines make use of only a subset of this information. Neither the Bayesian model nor the baselines make use of the raw images themselves. 

We use multiple metrics to compare the performance of our Bayesian model to baselines. First, we assess the out-of-sample Pearson correlation with fraction of images in the tract which are classified flooded. For this task, the output from our Bayesian model that we use is $p(y=1|C=c)$, since this captures the fraction of images in a Census tract which are flooded. 

We also assess the AUC in predicting whether (1) a Census tract will have any \emph{classified} flooded images and (2) a Census tract will have any \emph{ground-truth} annotated flooded images. For both these tasks, the output from our Bayesian model that we use is $p(\text{at least one image in tract is flooded} | C=c)$, since the goal is to predict the existence of a single image (as opposed to the fraction of flooded images). 

On all of these metrics, our Bayesian model outperforms all baselines (Table \ref{tab:baselines}).

\begin{figure}
    \includegraphics[width=0.48\textwidth]{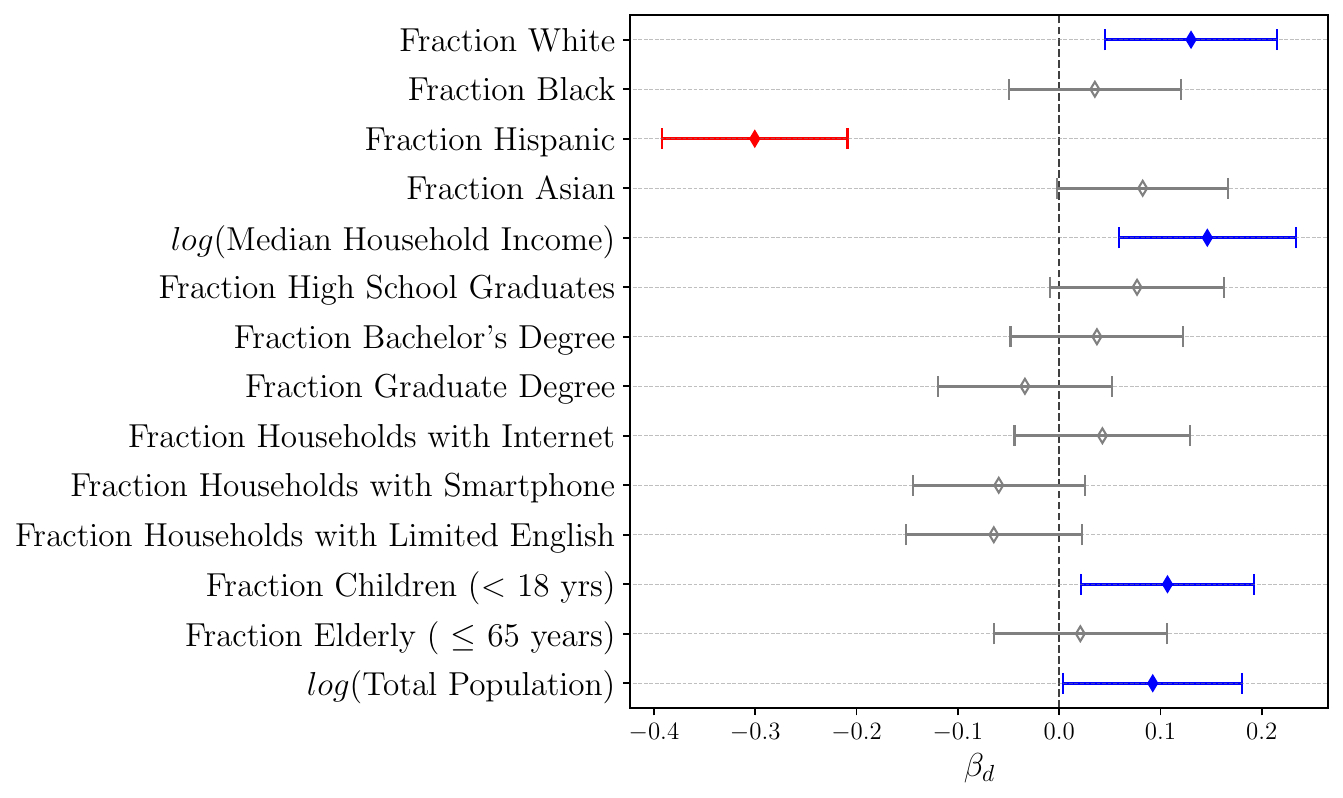}
    \caption{Demographic coefficients for a risk-adjusted regression where we control for whether a tract has at least one confirmed flooded image as our measure of flood risk (as opposed to controlling for $ \floodrisk $, as in our main results). Coefficients remain similar (although the statistical significance of some coefficients changes).}
    \label{fig:bias-confirmed-flooding-only}
\end{figure}

\end{document}